\pdfoutput=1

\documentclass[11pt]{article}

\usepackage{ACL2023}

\usepackage{times}
\usepackage{latexsym}

\usepackage[T1]{fontenc}

\usepackage[utf8]{inputenc}

\usepackage{microtype}

\usepackage{inconsolata}
\usepackage{graphicx}
\usepackage{subfigure}
\usepackage{booktabs} 

\usepackage{amsfonts}
\usepackage{stfloats}
\usepackage{ulem}

\usepackage{amsmath}

\usepackage{adjustbox}
\usepackage{multirow}
\usepackage{array}
\usepackage{float}
\usepackage{caption}
\usepackage{balance}
\usepackage{flushend}

\usepackage{xurl}

\usepackage{listings}
\usepackage{color}

\definecolor{dkgreen}{rgb}{0,0.6,0}
\definecolor{gray}{rgb}{0.5,0.5,0.5}
\definecolor{mauve}{rgb}{0.58,0,0.82}

\usepackage[group-separator={,}]{siunitx}

\title{Chinese CLIP: Contrastive Vision-Language Pretraining in Chinese}

\author{An Yang$^{\ast1}$, Junshu Pan$^{\ast1,2\dagger}$, Junyang Lin$^{\ast1}$, \\
\textbf{Rui Men${^1}$, Yichang Zhang${^1}$, Jingren Zhou${^1}$, Chang Zhou$^{1\clubsuit}$ } \\
${^1}$DAMO Academy, Alibaba Group \\ 
${^2}$Beihang University \\
\{ya235025, panjunshu.pjs, junyang.ljy, ericzhou.zc\}@alibaba-inc.com
}

\begin{document}
\maketitle
\newcommand\blfootnote[1]{%
\begingroup
\renewcommand\thefootnote{}\footnote{#1}%
\addtocounter{footnote}{-1}%
\endgroup
}
\blfootnote{$^{\ast}$Co-first authors.}
\blfootnote{$^{\clubsuit}$Corresponding author. }
\blfootnote{$^{\dagger}$Work done as an intern in DAMO Academy.}

\begin{abstract}
The tremendous success of vision-language foundation models has promoted the research and application of computer vision and multimodal representation leearning. 
However, it is still difficult to effectively transfer such foundation models to language-specific scenarios. 
In this work, we propose Chinese CLIP with the two-stage pretraining method which trains the model with locked-image tuning in the first stage and contrastive tuning in the second one. 
Specifically, we have developed $5$ Chinese CLIP models of multiple sizes, spanning from $77$ to $958$ million parameters, and we have pretrained them on a collected large-scale dataset of Chinese image-text pairs. 
Our comprehensive experiments demonstrate that Chinese CLIP can achieve the state-of-the-art performance on MUGE, Flickr30K-CN, and COCO-CN in the setups of zero-shot learning and finetuning, and it is able to achieve competitive performance in zero-shot image classification based on the evaluation on the ELEVATER benchmark. 
We have released our codes, models, and demos\footnote{Github: \url{https://github.com/OFA-Sys/Chinese-CLIP}; ModelScope: \url{https://www.modelscope.cn/models}}. 
\end{abstract}

\section{Introduction}

\begin{figure}[t] 
    \centering
    \includegraphics[width=1.0\linewidth]{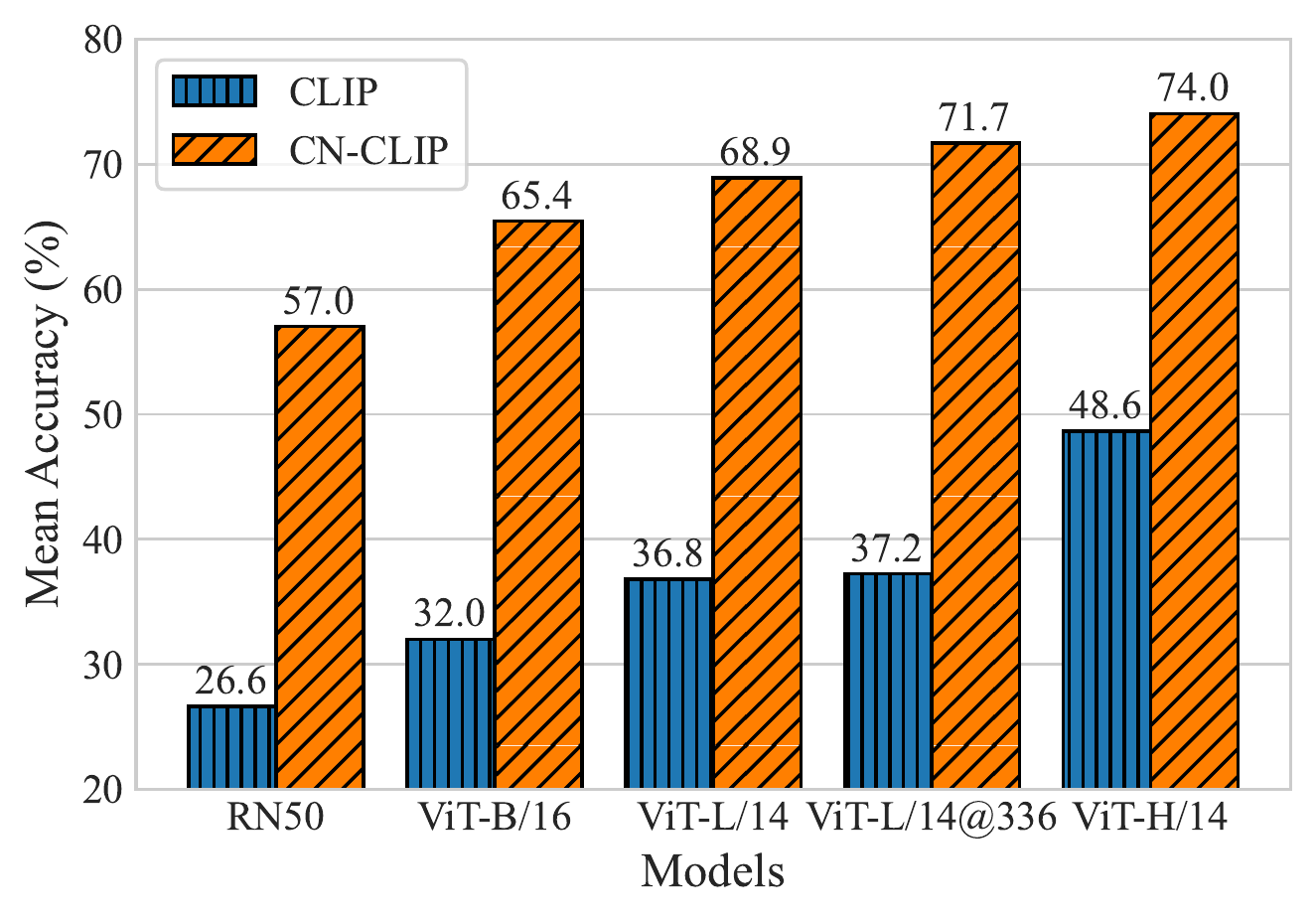}
    \caption{\textbf{Comparison of CLIP and Chinese CLIP models on the Chinese native retrieval benchmark MUGE.} On the benchmark based on the native data (which are mostly crawled from the language-native websites, in contrast with the translated data from websites of other countries.), CLIP performs far worse than our Chinese CLIP. Note that $\text{CLIP}\rm_{ViT\text{-}H/14}$ is not released, we use the model from OpenCLIP~\citep{openclip} instead.}
    \label{fig:comparison}
\end{figure}

Starting from the burst of pretraining in NLP, foundation models have attracted attention from multiple research communities. 
Foundation models that learn from large-scale unsupervised or weakly supervised data play as the basis of downstream models. 
A milestone of foundation models~\citep{foundation_model} in multimodal representation learning is CLIP~\cite{clip}. 
Different from the conventional generative pretraining, CLIP is a contrastive-learning-based model pretrained on a large-scale dataset of around $400$ million image-text pair data collected from the web. 
Despite the simplicity of the method, CLIP not only achieved outstanding performance in vision-language retrieval but more importantly played as a vision foundation model and demonstrated state-of-the-art performance in zero-shot image classification across a series of datasets. 
CLIP which builds a connection between vision and language has been transforming the research in both multimodal representation learning and computer vision. 

Be that as it may, it is difficult to efficiently transfer a cross-modal pretrained model to another language for several causes. 
First, learning to model the distribution of language-native vision and language data is significant for the transfer. 
Though CLIP performs as a strong foundation model in most scenarios, we find that it is hard for CLIP with machine translation to perform well on the Chinese-native cross-modal retrieval benchmark.
Figure~\ref{fig:comparison} demonstrates large performance gaps between the original CLIP and our Chinese CLIP at all model scales. 
We assume that it is crucial for both encoders to learn from the language-native images and texts. 
Second, the performance of previous methods for Chinese multimodal pretraining has been inhibited by several factors. Pretraining from scratch requires collecting a large-scale quality language-specific image text pair dataset similar to Web Image Text (WIT) for OpenAI CLIP~\citep{wenlan, r2d2}. Though the fast transfer of CLIP to Chinese data can be realized by using CLIP initialization and Locked-Image Tuning~\citep{lit}, the vision encoder still cannot learn the information of images from the language-specific domains~\citep{wukong}. 

\begin{figure}[t] 
    \centering
    \includegraphics[width=1.0\linewidth]{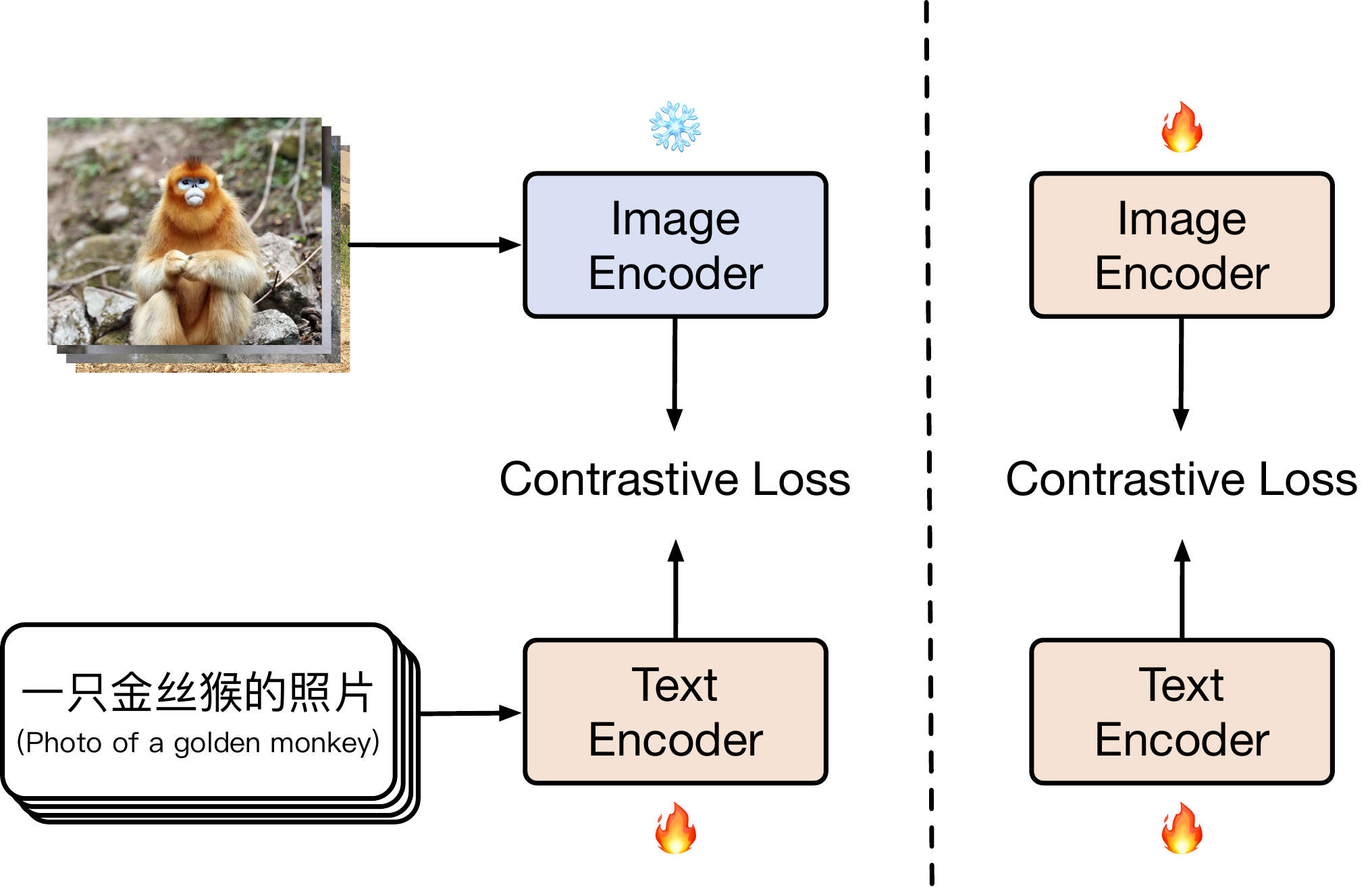}
    \caption{\textbf{An illustration of pretraining Chinese CLIP.} To leverage the advantages of the existing pretrained models, we initialize the image encoder with the OpenAI CLIP models, and the text encoder with the Chinese RoBERTa models. In Stage 1, we freeze the weights of the image encoder to avoid weight optimization, and in Stage 2, we unfreeze it and optimize both encoders. }
    \label{fig:model}
\end{figure}

Therefore, we propose Chinese CLIP, a language-specific vision-language foundation model pretrained on the publicly available Chinese image-text pair data.  
Additionally, we still use the same architecture as OpenAI CLIP.  
To realize efficient transfer of cross-modal foundation model to Chinese data, we develop a two-stage pretraining method, which is also adaptive to other vision-language foundation models, e.g., ALIGN, Florence, etc. 
Here in this work, we use CLIP as an example. 
To be specific, we first initialize both encoders with pretrained models, namely vision encoders from CLIP and text encoders from RoBERTa-wwm-Chinese~\citep{wwm}. In Stage 1, we freeze the image encoder and only optimize the text encoder with LiT, and in Stage 2, we train both encoders with contrastive tuning~\ref{fig:model}. 
In this way, the new model can inherit from the foundation models through initialization and LiT, and effectively transfer to language-specific data through contrastive tuning. 

We evaluate Chinese CLIP on $3$ Chinese cross-modal retrieval datasets, including MUGE\footnote{\url{https://tianchi.aliyun.com/muge}}, Flickr30K-CN~\citep{flickr30k-cn}, and COCO-CN~\citep{coco-cn}. Experimental results demonstrate that both the large-size and huge-size Chinese CLIP reach state-of-the-art performance on the $3$ datasets in the setups of both zero-shot learning and finetuning. 
Additionally, we evaluate the capability of zero-shot image classification on the track ``Image Classification in the Wild'' of the ELEVATER benchmark~\citep{elevater}. 
On the classification datasets, Chinese CLIP demonstrates competitive performance in comparison with state-of-the-art methods and outperforms the Chinese baselines. 
Furthermore, we provide NVIDIA TensorRT and ONNX models for deployments, which run around $2$ to $10$ times faster than Pytorch models for inference.

In brief, our contributions are:
\begin{itemize}
    \item We propose Chinese CLIP, a simple implementation of CLIP pretrained on our collected large-scale Chinese image-text pair data, and we propose a two-stage pretraining method to achieve high pretraining efficiency and improved downstream performance.
    \item Chinese CLIP achieves state-of-the-art performance in cross-modal retrieval in the setups of zero-shot learning and finetuning, and competitive performance in zero-shot image classification. 
\end{itemize}

\section{Method}

CLIP~\citep{clip} based on simple vision-language contrastive pretraining on large-scale weakly supervised data is a significant foundation model in multimodal representation learning. It can transfer to cross-modal retrieval directly, and its image encoder can play as a vision backbone. 
In this work, we propose to build a language-specific CLIP model by pretraining a vision-language model on large-scale Chinese multimodal data. 
In the following, we provide the details of method design and implementation of our Chinese CLIP. 

\subsection{Data}
One key to CLIP's success should be the large-scale dataset for pretraining. 
Based on the experiments of a CLIP reimplementation~\citep{openclip}, scaling up data and lengthening the training process can consistently improve the model performance in zero-shot learning. 
This year, the most recent multimodal pretrained models Wukong~\citep{wukong} and R2D2~\citep{r2d2} were pretrained on a public dataset of $100$ million image-text pairs and an in-house dataset of $250$ million samples, where only a subset of $23$ million samples were released. 
For the facility in reimplementation, we aim at pretraining Chinese CLIP on as many publicly available data as possible, and thus we focus on collecting high-quality public datasets. 
We extract the Chinese data (with the mark ``zh'') from the latest LAION-5B~\cite{schuhmann2021laion}, and collect the data from the Wukong dataset. 
However, due to the problems of unavailable links, we can only collect around $108$ million samples and $72$ million samples from LAION-5B and Wukong respectively. 
We additionally add the translated data from the classic English multimodal datasets, including Visual Genome~\cite{vg} and MSCOCO~\citep{coco_cap}, where test sets are removed. 
Finally, we construct a dataset for Chinese multimodal pretraining with around $200$ million image-text pairs.\footnote{We add around $20$ million high-quality internal image-text pairs to provide more diversity. } 

Below illustrates the procedure for data preprocessing. 
For the part of data from LAION-5B, we remove the samples with CLIP scores lower than $0.26$ computed by mCLIP~\citep{mclip}. 
Besides, we remove the samples with captions containing words in our internal blacklist. 
The blacklist contains words related to advertising, image filenames, etc. 
We remove those samples that are too short (fewer than $5$ characters) or too long (more than $50$ characters). 
For the images, we resize them to the resolution of $224 \times 224$ for most cases and $336 \times 336$ for the ViT-L/14@336px. 

\subsection{Pretraining Method}
\label{subsec:pretraining_method}

There are multiple design choices for pretraining the Chinese CLIP models. 
One of the simplest methods should be pretraining from scratch, where both the image and text encoders are randomly initialized. 
However, we assume that its performance will be limited by the quantity and quality of the current pretraining data. 
To leverage the advantages of existent pretrained models, we initialize the models with weights from the pretrained checkpoints from the official release of CLIP~\footnote{\url{https://github.com/openai/CLIP} License: MIT.} for the image encoder, and RoBERTa-wwm-ext and RBT3~\footnote{\url{https://github.com/ymcui/Chinese-BERT-wwm} License: Apache 2.0} for the text encoder. 
To adapt the model to the introduced pretraining data, it is available to pretrain it with ``contrastive tuning'', similar to the way to transfer CLIP to downstream retrieval data. 
In comparison with contrastive tuning, Locked-image Tuning (LiT)~\citep{lit} demonstrated improved performance in downstream transfer. 

In this work, we propose a two-stage pretraining method, as shown in Figure~\ref{fig:model}. 
The core idea is to first utilize LiT to enable the text encoder to read out high-quality representations from the foundation vision model from OpenAI CLIP, and then transfer the whole model to the domain of the introduced pretraining data. 
It is not sufficient to pretrain Chinese CLIP with solely LiT, as the image encoder should learn the information of the images of the Chinese datasets and model the distribution of such data. 
Before the two-stage pretraining, we first initialize both encoders with pretrained models. 
In Stage 1, we ``lock'' the image encoder by freezing its parameters during pretraining. 
We only pretrain the text encoder for vision-language alignment, based on the assumption that the vision backbone with pretrained weights is already a powerful vision foundation model~\citep{lit, wukong}. 
We pretrain it until there is no salient performance improvement in downstream tasks, even if we prolong the pretraining progress. 
Then we switch to Stage 2, where we ``unlock'' the image encoder by enabling its optimization. 
In Stage 2, we continue pretraining without any parameter frozen, so that the image encoder can learn to model the distribution of the image data from Chinese websites. 
In the ablation study, we discuss the influence of the initialization of the pretrained checkpoints and pretraining methods on the downstream performance. 
Experimental results show that the two-stage pretraining method can outperform either pretraining from scratch or directly finetuning from the pretrained models.

\begin{table*}[t]
\center
\small
\vskip 0.15in
\begin{adjustbox}{max width=1.\textwidth}
\begin{tabular}{@{\extracolsep{\fill}}lccccccccc}
\toprule
  Tasks
  &\multicolumn{4}{c}{Zero-shot}
  &\multicolumn{4}{c}{Finetuning}
  \\
\midrule
  Metrics & R@1 & R@5 & R@10 & MR & R@1 & R@5 & R@10 & MR
  \\
\midrule
    \multicolumn{9}{l}{\textit{Tiny}-size Model} \\
    $\text{CN-CLIP}\rm_{RN50}$
    & \textbf{42.6}	& \textbf{68.6}	& \textbf{77.9}	& \textbf{63.0}	& \textbf{48.6}	& \textbf{75.1}	& \textbf{84.0}	& \textbf{69.2}
    \\
\midrule
    \multicolumn{9}{l}{\textit{Base}-size Models} \\
    $\text{Wukong}\rm_{ViT\text{-}B/32}$
    & 33.4	& 59.3	& 69.7	& 54.1	& 39.2	& 66.9	& 77.4	& 61.2
    \\
    $\text{R2D2}\rm_{ViT\text{-}B}$
    & -	& -	& -	& -	& 47.4	& 75.1	& 83.5	& 68.7
    \\
    $\text{CN-CLIP}\rm_{ViT\text{-}B/16}$
    & \textbf{52.1}	& \textbf{76.7}	& \textbf{84.4}	& \textbf{71.1}	& \textbf{58.4}	& \textbf{83.6}	& \textbf{90.0}	& \textbf{77.4}
    \\
\midrule
    \multicolumn{9}{l}{\textit{Large}-size Models} \\
    $\text{Wukong}\rm_{ViT\text{-}L/14}$
    & 42.7	& 69.0	& 78.0	& 63.2	& 52.7	& 77.9	& 85.6	& 72.1
    \\
    $\text{R2D2}\rm_{ViT\text{-}L/14}$
    & 49.5 	& 75.7 	& 83.2	& 69.5	& 60.1 	& 82.9 	& 89.4 	& 77.5
    \\
    $\text{CN-CLIP}\rm_{ViT\text{-}L/14}$
    & 56.3	& 79.8	& 86.2	& 74.1	& 63.3	& 85.6	& 91.3	& 80.1
    \\
    $\text{CN-CLIP}\rm_{ViT\text{-}L/14@336px}$
    & \textbf{59.0}	& \textbf{81.4}	& \textbf{87.8}	& \textbf{76.1}	& \textbf{65.3}	& \textbf{86.7}	& \textbf{92.1}	& \textbf{81.3}
    \\
\midrule
    \multicolumn{9}{l}{\textit{Huge}-size Models} \\
    $\text{CN-CLIP}\rm_{ViT\text{-}H/14}$
    & \textbf{63.0}	& \textbf{84.1}	& \textbf{89.2}	& \textbf{78.8}	& \textbf{68.9}	& \textbf{88.7}	& \textbf{93.1}	& \textbf{83.6}
    \\
\bottomrule
\end{tabular}
\end{adjustbox}
\caption{Experimental results on MUGE-Retrieval. We report the performance of both baselines and Chinese CLIP models on text-to-image retrieval and image-to-text retrieval in the setups of zero-shot evaluation and finetuning. }
\label{tb:muge}
\end{table*}

\begin{table*}[t]
\center
\small
\vskip 0.15in
\begin{adjustbox}{max width=1.\textwidth}
\begin{tabular}{@{\extracolsep{\fill}}lccccccccccccc}
\toprule
  Tasks
  &\multicolumn{6}{c}{Text-to-Image}
  &\multicolumn{6}{c}{Image-to-Text}
  \\
\midrule
  Setups
  &\multicolumn{3}{c}{Zero-shot}
  &\multicolumn{3}{c}{Finetuning}
  &\multicolumn{3}{c}{Zero-shot}
  &\multicolumn{3}{c}{Finetuning}
  \\
\midrule
  Metrics & R@1 & R@5 & R@10 & R@1 & R@5 & R@10 & R@1 & R@5 & R@10 & R@1 & R@5 & R@10
  \\
\midrule
    \multicolumn{13}{l}{\textit{Tiny}-size Model} \\
    $\text{CN-CLIP}\rm_{RN50}$
    & \textbf{48.8}	& \textbf{76.0}	& \textbf{84.6}	& \textbf{66.7} & \textbf{89.4} & \textbf{94.1} & \textbf{60.0}	& \textbf{85.9}	& \textbf{92.0}	& \textbf{84.2}	& \textbf{96.7}	& \textbf{98.0}
    \\
\midrule
    \multicolumn{13}{l}{\textit{Base}-size Models} \\
    $\text{Wukong}\rm_{ViT\text{-}B/32}$
    & 45.7	& 73.8	& 82.2	& 67.6	& 89.6	& 94.2	& 66.2	& 88.7	& 94.3	& 83.9	& 97.6	& 99.0
    \\
    $\text{R2D2}\rm_{ViT\text{-}B}$
    & -	& -	& -	& 78.3	& 94.6	& 97.0	& -	& -	& -	& 92.6	& \textbf{99.1}	& \textbf{99.8}
    \\
    $\text{CN-CLIP}\rm_{ViT\text{-}B/16}$
    & \textbf{62.7}	& \textbf{86.9}	& \textbf{92.8}	& \textbf{79.1}	& \textbf{94.8}	& \textbf{97.4}	& \textbf{74.6}	& \textbf{93.5}	& \textbf{97.1}	& \textbf{93.5}	& 99.0	& 99.5
    \\
\midrule
    \multicolumn{9}{l}{\textit{Large}-size Models} \\
    $\text{Wukong}\rm_{ViT\text{-}L/14}$
    & 51.7	& 78.9	& 86.3	& 77.4 	& 94.5 	& 97.0	& 76.1	& 94.8	& 97.5	& 92.7 	& 99.1 	& 99.6
    \\
    $\text{R2D2}\rm_{ViT\text{-}L/14}$
    & 60.9	& 86.8	& 92.7	& \textbf{84.4} 	& 96.7 	& 98.4	& 77.6	& 96.7	& \textbf{98.9}	& 95.6 	& \textbf{99.8}	& \textbf{100.0}
    \\
    $\text{CN-CLIP}\rm_{ViT\text{-}L/14}$
    & 68.0	& 89.7	& 94.4	& 82.7	& 96.7	& 98.6	& 80.2	& 96.6	& 98.2	& 96.1	& 99.5	& 99.9
    \\
    $\text{CN-CLIP}\rm_{ViT\text{-}L/14@336px}$
    & \textbf{69.0}	& \textbf{90.7}	& \textbf{95.4}	& \textbf{84.4}	& \textbf{97.1}	& \textbf{98.7}	& \textbf{83.3}	& \textbf{97.2}	& 98.5	& \textbf{96.6}	& \textbf{99.8}	& \textbf{100.0}
    \\
\midrule
    \multicolumn{9}{l}{\textit{Huge}-size Models} \\
    $\text{CN-CLIP}\rm_{ViT\text{-}H/14}$
    & \textbf{71.2}	& \textbf{91.4}	& \textbf{95.5}	& \textbf{83.8}	& \textbf{96.9}	& \textbf{98.6}	& \textbf{81.6}	& \textbf{97.5}	& \textbf{98.8}	& \textbf{95.3}	& \textbf{99.7}	& \textbf{100.0}
    \\
\bottomrule
\end{tabular}
\end{adjustbox}
\caption{Experimental results on Flickr30K-CN. We report the performance of both baselines and Chinese CLIP models on text-to-image retrieval and image-to-text retrieval in the setups of zero-shot evaluation and finetuning. }
\label{tb:flickr}
\end{table*}

\begin{table*}[t]
\center
\small
\vskip 0.15in
\begin{adjustbox}{max width=1.\textwidth}
\begin{tabular}{@{\extracolsep{\fill}}lccccccccccccc}
\toprule
  Tasks
  &\multicolumn{6}{c}{Text-to-Image}
  &\multicolumn{6}{c}{Image-to-Text}
  \\
\midrule
  Setups
  &\multicolumn{3}{c}{Zero-shot}
  &\multicolumn{3}{c}{Finetuning}
  &\multicolumn{3}{c}{Zero-shot}
  &\multicolumn{3}{c}{Finetuning}
  \\
\midrule
  Metrics & R@1 & R@5 & R@10 & R@1 & R@5 & R@10 & R@1 & R@5 & R@10 & R@1 & R@5 & R@10 
  \\
\midrule
    \multicolumn{13}{l}{\textit{Tiny}-size Model} \\
    $\text{CN-CLIP}\rm_{RN50}$
    & \color{gray}{48.1}	& \color{gray}{81.3}	& \color{gray}{90.5}	& \textbf{66.8}	& \textbf{91.1}	& \textbf{97.0}	& \color{gray}{51.6}	& \color{gray}{81.2}	& \color{gray}{90.5}	& \textbf{68.4}	& \textbf{93.3}	& \textbf{97.8}
    \\
\midrule
    \multicolumn{13}{l}{\textit{Base}-size Models} \\
    $\text{Wukong}\rm_{ViT\text{-}B/32}$
    & 49.2	& 79.4	& 87.9	& 67.0	& 91.4	& 96.7	& 48.3	& 77.8	& 88.8	& 65.8	& 90.3	& 96.6
    \\
    $\text{R2D2}\rm_{ViT\text{-}B}$
    & -	& -	& -	& 75.1	& 94.2	& 98.1	& -	& -	& -	& 76.1	& 95.3	& 98.5
    \\
    $\text{CN-CLIP}\rm_{ViT\text{-}B/16}$
    & \color{gray}{62.2}	& \color{gray}{86.6}	& \color{gray}{94.9}	& \textbf{77.0}	& \textbf{97.1}	& \textbf{99.0}	& \color{gray}{57.0}	& \color{gray}{84.1}	& \color{gray}{93.6}	& \textbf{77.4}	& \textbf{96.2}	& \textbf{98.9}
    \\
\midrule
    \multicolumn{9}{l}{\textit{Large}-size Models} \\
    $\text{Wukong}\rm_{ViT\text{-}L/14}$
    & 53.4	& 80.2	& 90.1	& 74.0	& 94.4	& 98.1	& 55.2	& 81.0	& 90.6	& 73.3	& 94.0	& 98.0
    \\
    $\text{R2D2}\rm_{ViT\text{-}L/14}$
    & 56.4	& 85.0	& 93.1	& 79.1	& 96.5	& 98.9	& 63.3	& 89.3	& 95.7	& 79.3	& 97.1	& 98.7
    \\
    $\text{CN-CLIP}\rm_{ViT\text{-}L/14}$
     & \color{gray}{64.0}	  & \color{gray}{89.2}	 & \color{gray}{94.4}	 & 78.9	 & 96.3	 & 99.0	 & \color{gray}{60.4}	 & \color{gray}{84.2}	 & \color{gray}{92.9}	 & 80.2	 & 96.7	 & \textbf{99.2}
    \\
    $\text{CN-CLIP}\rm_{ViT\text{-}L/14@336px}$
    & \color{gray}{64.7}	& \color{gray}{89.6}	& \color{gray}{94.6}	& \textbf{80.1}	& \textbf{96.7}	& \textbf{99.2} & \color{gray}{63.4}	& \color{gray}{87.2}	& \color{gray}{94.4}	& \textbf{81.2}	& \textbf{97.2}	& 99.1
    \\
\midrule
    \multicolumn{9}{l}{\textit{Huge}-size Models} \\
    $\text{CN-CLIP}\rm_{ViT\text{-}H/14}$
    & \color{gray}{69.2}	& \color{gray}{89.9}	& \color{gray}{96.1}	& \textbf{81.5}	& \textbf{96.9}	& \textbf{99.1}	& \color{gray}{63.0}	& \color{gray}{86.6}	& \color{gray}{92.9}	& \textbf{83.5}	& \textbf{97.3}	& \textbf{99.2}
    \\    
\bottomrule
\end{tabular}
\end{adjustbox}
\caption{Experimental results on COCO-CN. We report the performance of both baselines and Chinese CLIP models on text-to-image retrieval and image-to-text retrieval in the setups of zero-shot evaluation and finetuning. Since machine translated COCO is included in our pretraining dataset, here the numbers of Chinese CLIP zero-shot performances are shown in gray.}
\label{tb:coco}
\end{table*}

\section{Evaluation}
To comprehensively probe the effects of Chinese CLIP, we follow the conventional practice that we first evaluate its basic capabilities of cross-modal retrieval, i.e. text-to-image retrieval and image-to-text retrieval, in different domains, including e-commerce and the general domain. 
Additionally, as the contrastive-learning-based pretraining builds a foundation vision model that is semantically connected with natural language, we follow \citet{clip} and evaluate its capabilities of zero-shot classification. Specifically, we validate Chinese CLIP on the classification datasets of the ELEVATER benchmark~\citep{elevater}, which is known as ``Image Classification in the Wild (ICinW)''. 

\subsection{Cross-modal Retrieval}

\subsubsection{Datasets and Metrics}
We validate Chinese CLIP on $3$ cross-modal retrieval datasets, namely MUGE-Retrieval, Flickr30K-CN~\cite{flickr30k-cn}, and COCO-CN~\cite{coco-cn}. 
MUGE-Retrieval is an image-text retrieval dataset, where data are extracted from Chinese E-commerce websites. 
Flickr30K-CN and COCO-CN are built from the classical datasets Flickr30K and MSCOCO-1K whose texts are translated into Chinese. 
Our evaluation includes setups of zero-shot learning and finetuning. 
For zero-shot learning, we use Chinese CLIP models to compute the similarity scores between images and texts and return the top-$K$ most similar candidates. For finetuning, we finetune the Chinese CLIP models for cross-modal retrieval with contrastive tuning. The evaluation is the same as that in zero-shot learning. 
The evaluation metrics are Recall@$K$, where $K = \{1, 5, 10\}$, and Mean Recall (MR, i.e., the average of Recall@$K$). 
For comparison, we choose the base-size and large-size Wukong and R2D2 as the baselines, which are the previous SOTA models in Chinese multimodal representation learning. 
Following these baselines, we report validation performance on MUGE and test performance on Flickr30K-CN and COCO-CN.
Note that in the setup of finetuning, R2D2\footnote{Since the original paper of R2D2 does not provide the patch size of their base-size model, here we denote the model as $\text{R2D2}\rm_{ViT\text{-}B}$.} is essentially an end-to-end model of retrieval and ranking. 

\subsubsection{Results}
\label{subsubsec:results}
Table~\ref{tb:muge} reports the model performance on MUGE-Retrieval. For the base-size model, $\text{CN-CLIP}\rm_{ViT\text{-}B/16}$ outperforms the baselines on all metrics and in both setups of zero-shot learning and finetuning. 
Specifically, for the base-size models, $\text{CN-CLIP}\rm_{ViT\text{-}B/16}$ surpasses $\text{Wukong}\rm_{ViT\text{-}B/32}$ by $17.0$ MR in zero-shot learning and surpasses $\text{R2D2}\rm_{ViT\text{-}B}$ by $8.7$ MR in finetuning. 
Besides, the tiny model $\text{CN-CLIP}\rm_{RN50}$ can outperform the base-size $\text{Wukong}\rm_{ViT\text{-}B/32}$ by $8.9$ MR in zero-shot learning and $8.0$ MR in finetuning. 

For the large-size models, $\text{CN-CLIP}\rm_{ViT\text{-}L/14}$ can outperform both baselines in all metrics and $\text{CN-CLIP}\rm_{ViT\text{-}L/14@336px}$ pretrained on images of a larger resolution can achieve the state-of-the-art performance. $\text{CN-CLIP}\rm_{ViT\text{-}L/14@336px}$ outperforms $\text{R2D2}\rm_{ViT\text{-}L/14}$ by $6.6$ MR in zero-shot learning and $3.8$ MR in finetuning. 
When scaling to $\text{CN-CLIP}\rm_{ViT\text{-}H/14}$, the performance is further improved. Compared with the best large-size model $\text{CN-CLIP}\rm_{ViT\text{-}L/14@336px}$, $\text{CN-CLIP}\rm_{ViT\text{-}H/14}$ surpasses it by $2.7$ MR in zero-shot learning and $2.3$ MR in finetuning. 

Table~\ref{tb:flickr} and \ref{tb:coco} report the model performance on Flickr30K-CN and COCO-CN. 
We focus on the evaluation of R@1. 
In both datasets, $\text{CN-CLIP}$ achieves better performance than the baselines. 
For the base-size models, in the setup of zero-shot learning of Flickr30K-CN, $\text{CN-CLIP}\rm_{ViT\text{-}B/16}$ surpasses $\text{Wukong}\rm_{ViT\text{-}B/32}$ by $17.0$ R@1 in text-to-image retrieval and $8.4$ R@1 in image-to-text retrieval, and in the finetuning setup, $\text{CN-CLIP}\rm_{ViT\text{-}B/16}$ surpasses $\text{R2D2}\rm_{ViT\text{-}B}$ by $0.8$ R@1 in image retrieval and $0.9$ R@1 in text retrieval. 
Similarly, in the finetuning setup of COCO-CN, $\text{CN-CLIP}\rm_{ViT\text{-}B/16}$ surpasses $\text{R2D2}\rm_{ViT\text{-}B}$ by $1.9$ R@1 in image retrieval and $1.3$ R@1 in text retrieval. For the tiny-size $\text{CN-CLIP}\rm_{RN50}$, it again achieves or surpasses the performance of $\text{Wukong}\rm_{ViT\text{-}B/32}$ in several metrics of Flickr30K-CN and COCO-CN. Specifically, $\text{CN-CLIP}\rm_{RN50}$ surpasses $\text{Wukong}\rm_{ViT\text{-}B/32}$ by $3.1$ R@1 in the zero-shot learning of Flickr30K-CN image retrieval and by $2.6$ R@1 in the finetuning of COCO-CN text retrieval.

For the large-size models, in the zero-shot setup of Flickr30K-CN, $\text{CN-CLIP}\rm_{ViT\text{-}L/14}$ surpasses $\text{Wukong}\rm_{ViT\text{-}L/14}$ by $16.3$ R@1 in text-to-image retrieval and $4.1$ R@1 in image-to-text retrieval. $\text{CN-CLIP}\rm_{ViT\text{-}L/14@336px}$ further improves over $\text{CN-CLIP}\rm_{ViT\text{-}L/14}$ by $1.0$ R@1 in image retrieval and $3.1$ R@1 in text retrieval. In the finetuning setup, $\text{CN-CLIP}\rm_{ViT\text{-}L/14}$ surpasses $\text{R2D2}\rm_{ViT\text{-}L/14}$ by $0.5$ R@1 in text retrieval. $\text{CN-CLIP}\rm_{ViT\text{-}L/14@336px}$ achieves equal performance with $\text{R2D2}\rm_{ViT\text{-}L/14}$ in image retrieval and surpasses it by $1.0$ R@1 in text retrieval. Similarly, in the finetuning setup of COCO-CN, $\text{CN-CLIP}\rm_{ViT\text{-}L/14}$ surpasses $\text{R2D2}\rm_{ViT\text{-}L/14}$ by $0.9$ R@1 in text retrieval. $\text{CN-CLIP}\rm_{ViT\text{-}L/14@336px}$ further surpasses $\text{R2D2}\rm_{ViT\text{-}L/14}$ by $1.0$ in image retrieval and $1.9$ R@1 in text retrieval. 

On Flickr30K-CN and COCO-CN, scaling from $\text{CN-CLIP}\rm_{ViT\text{-}L/14}$ to $\text{CN-CLIP}\rm_{ViT\text{-}H/14}$ improves the performance in almost all the metrics. Specifically, in the zero-shot setup of Flickr30K-CN, $\text{CN-CLIP}\rm_{ViT\text{-}H/14}$ surpasses $\text{CN-CLIP}\rm_{ViT\text{-}L/14}$ by $3.2$ R@1 in image retrieval and $1.4$ R@1 in text retrieval. Moreover, in the finetuning setup of COCO-CN, $\text{CN-CLIP}\rm_{ViT\text{-}H/14}$ even surpasses $\text{CN-CLIP}\rm_{ViT\text{-}L/14@336px}$ with larger image resolution by $1.4$ R@1 in image retrieval and $2.3$ R@1 in text retrieval. 
We also compare our $\text{CN-CLIP}\rm_{ViT\text{-}H/14}$ with a huge CLIP-like model, T-Bletchley\footnote{\url{https://www.microsoft.com/en-us/research/blog/turing-bletchley-a-universal-image-language-representation-model-by-microsoft/}}. This model has $2.5$ billion parameters and is pretrained on billions of multilingual image-caption pairs. In the finetuning setup of COCO-CN, with smaller sizes of model parameters and pretrain dataset, $\text{CN-CLIP}\rm_{ViT\text{-}H/14}$ still surpasses T-Bletchley by $3.9$ MR.

\begin{figure*}
    \centering
    \subfigure[]{
        \begin{minipage}[t]{0.66\columnwidth}
            \label{fig:MUGE IR}
            \includegraphics[width=\linewidth]{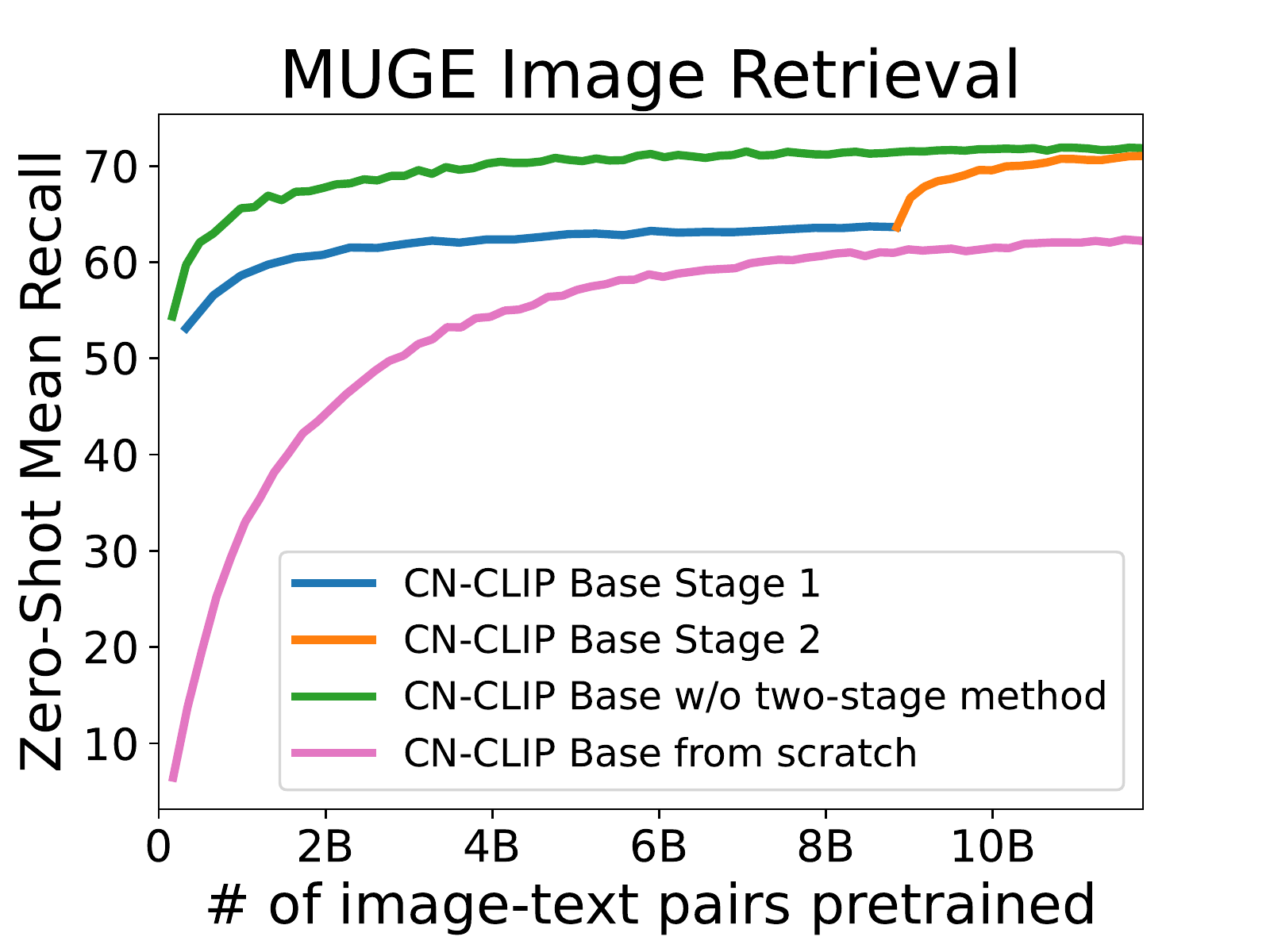}
        \end{minipage}
    }
    \subfigure[]{
        \begin{minipage}[t]{0.66\columnwidth}
            \label{fig:Flickr30k-CN IR}
            \includegraphics[width=\linewidth]{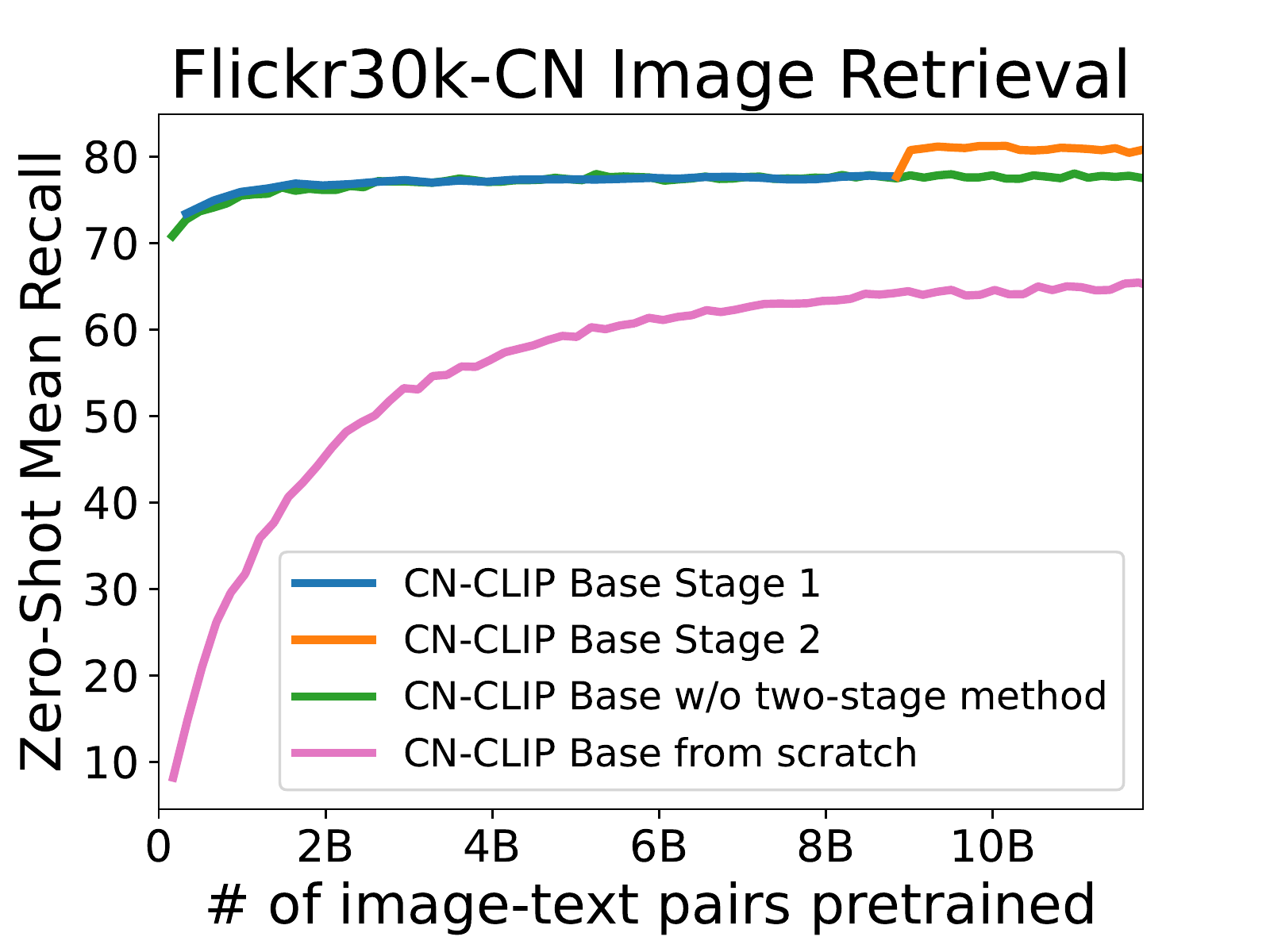}
        \end{minipage}
    }
    \subfigure[]{
        \begin{minipage}[t]{0.66\columnwidth}
            \label{fig:Flickr30k-CN TR}
            \includegraphics[width=\linewidth]{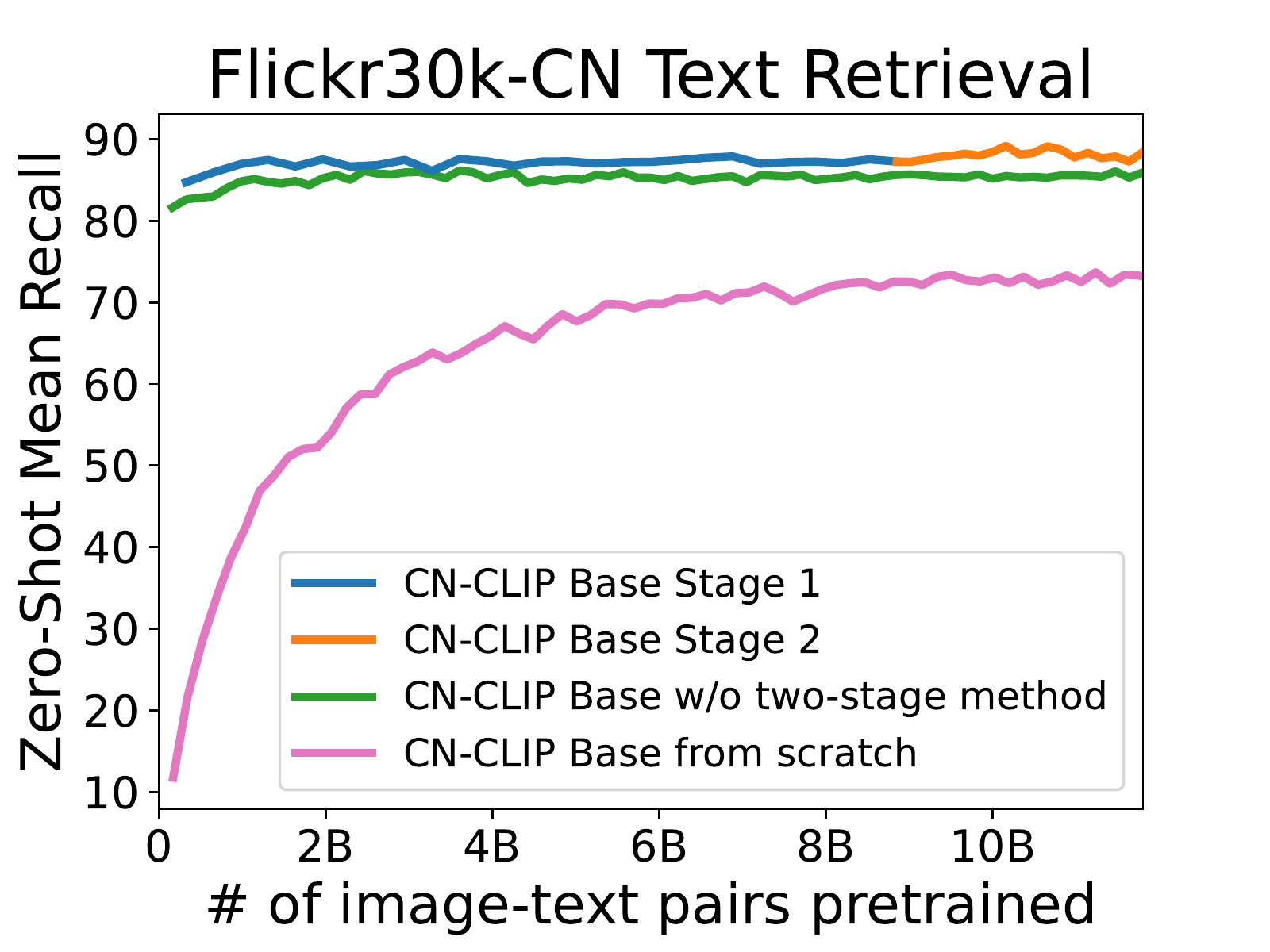}
        \end{minipage}
    }
    \subfigure[]{
        \begin{minipage}[t]{0.66\columnwidth}
            \label{fig:COCO-CN IR}
            \includegraphics[width=\linewidth]{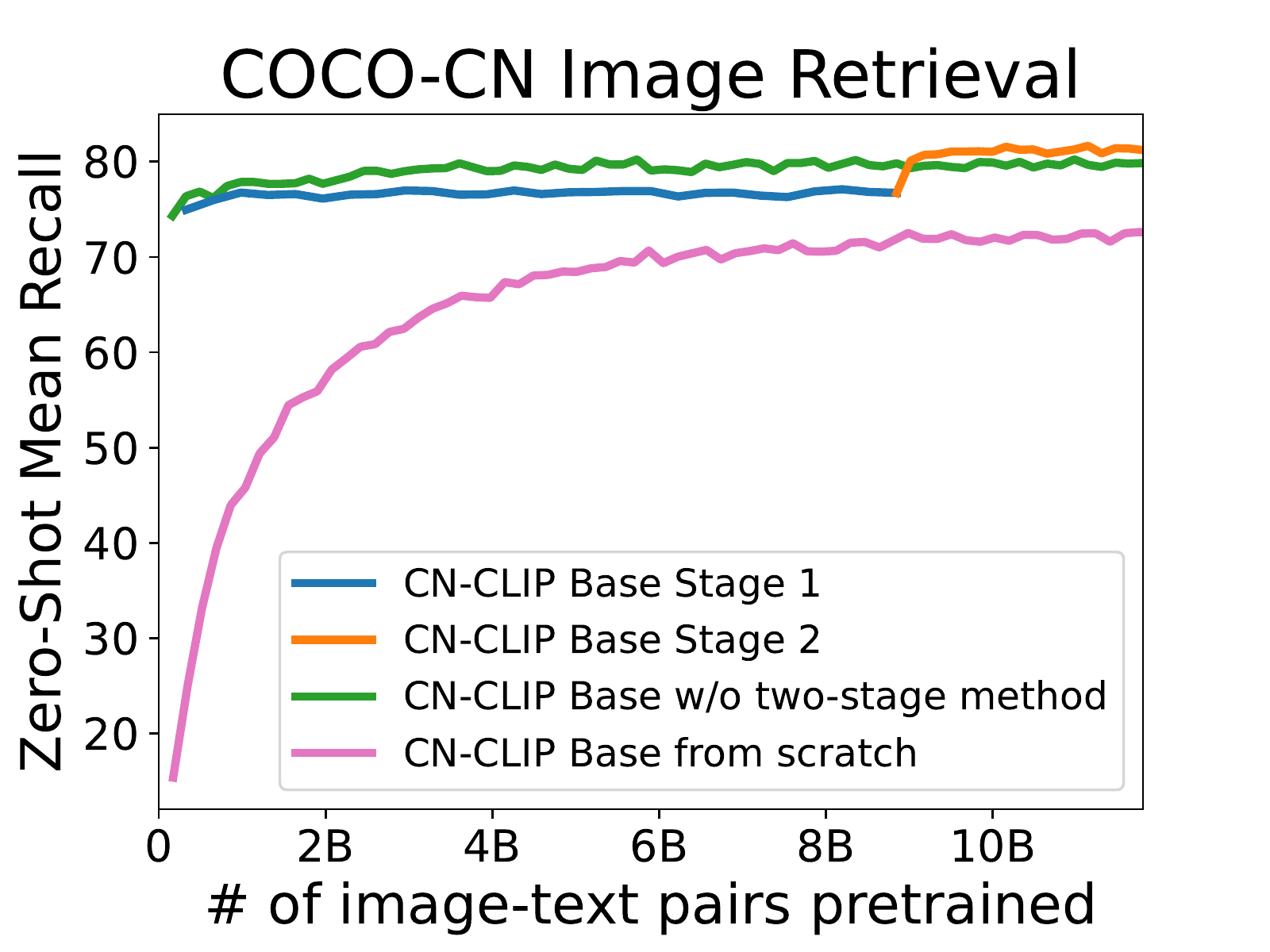}
        \end{minipage}
    }
    \subfigure[]{
        \begin{minipage}[t]{0.66\columnwidth}
            \label{fig:COCO-CN TR}
            \includegraphics[width=\linewidth]{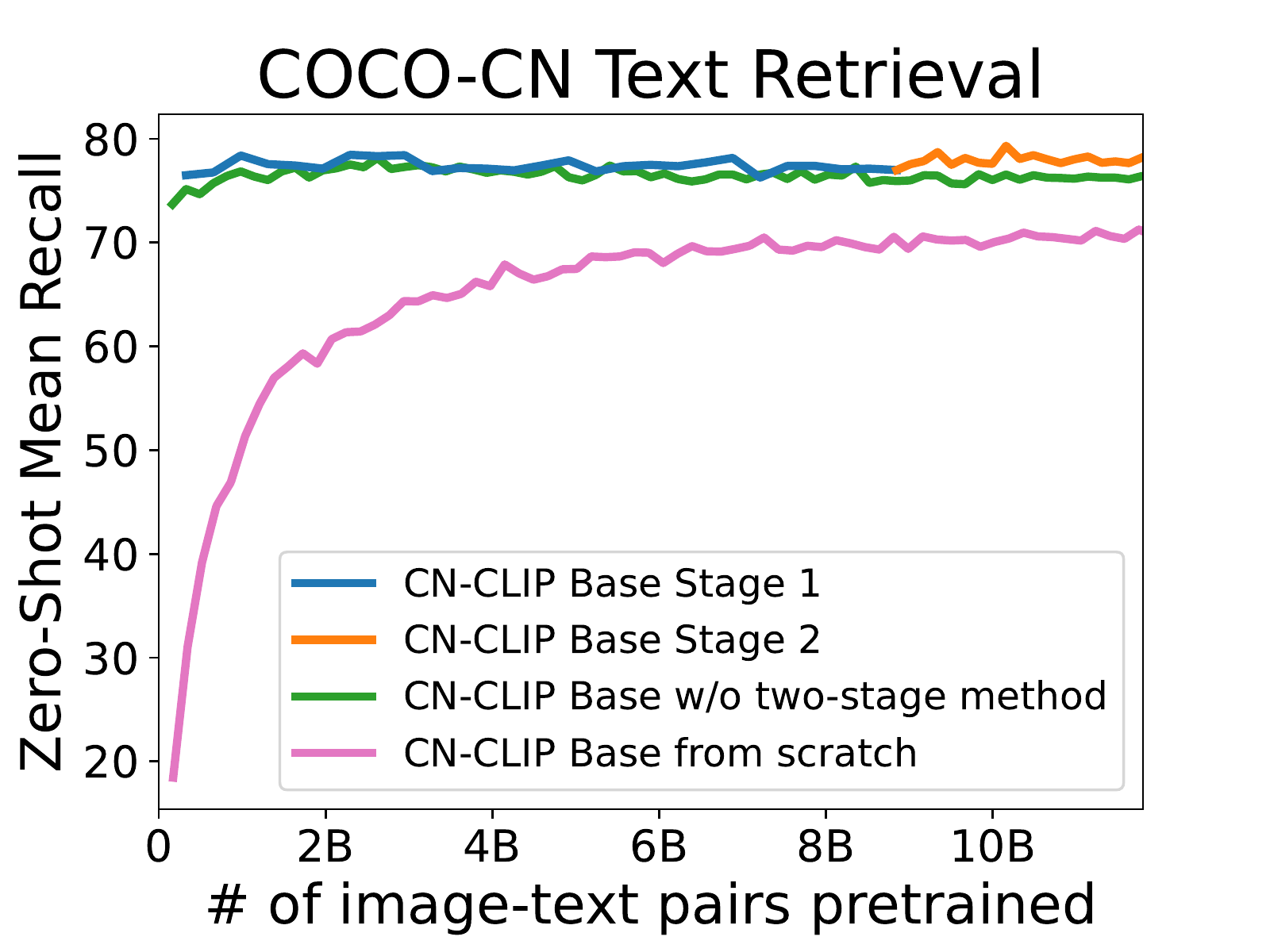}
        \end{minipage}
    }
    \caption{Comparison of base-size Chinese CLIP models with different training methods on MUGE, Flickr30k-CN and COCO-CN.}
    \label{fig:ablation Base MR}
\end{figure*}


\subsubsection{Ablation Study}
Here we provide an ablation study on our proposed two-stage training methods. 
To validate its significance and effectiveness, we design several setups for the ablation study. 
Our experiments are conducted on $\text{CN-CLIP}\rm_{ViT\text{-}B/16}$. 
To evaluate the influence of initialization, we pretrain a model from scratch, and to examine the influence of LiT, we pretrain a model without freezing the image encoder. 
For a better demonstration, we report the curves of model performance of zero-shot retrieval on different datasets in terms of pretraining progress, indicated by the number of processed samples. 

Figure~\ref{fig:ablation Base MR} shows the performance on different tasks, namely MUGE text-to-image retrieval, text-to-image and image-to-text retrieval on Flicrk30K-CN and COCO-CN. 
In comparison with pretraining with pretrained model initialization, pretraining from scratch performs much worse though it shows consistent performance improvement in terms of pretraining progress. 
As to the importance of LiT, we observe different phenomena on different datasets. On MUGE, a dataset of samples originally collected from Chinese websites, we find that pretraining without LiT might be the best solution though its performance gap with two-stage pretraining is quite small. 
However, on the other datasets, i.e., Flickr30K-CN and COCO-CN, whose samples are translated from the English datasets, we find that our two-stage pretraining performs significantly better than pretraining without LiT. 
Furthermore, we observe a common phenomenon that in the two-stage pretraining, switching from Stage 1 to Stage 2 can effectively boost the model performance to a higher level. 
This reflects the importance of adapting the pretrained model to the data distribution of the Chinese multimodal data, especially those concerned with visual information. 

\subsection{Zero-shot Image Classification}
\begin{table*}[t]
\center
\small
\vskip 0.15in
\begin{adjustbox}{max width=1.\textwidth}
\begin{tabular}{@{\extracolsep{\fill}}lccccccccccc}
\toprule
    Model & CIFAR10 & CIFAR100 & DTD & EuroSAT & FER & FGVC & KITTI & MNIST & PC & VOC & INet\\
\midrule
    \multicolumn{8}{l}{\textit{Original benchmark}} \\
    DeCLIP  & 90.9 & 66.8 & 44.9 & 39.9 & 23.3 & 9.0 & 39.7 & 13.6 & 55.3  & 80.6 & 73.7\\
    GIT  & 88.5 & 61.1 & 42.9 & 43.4 & 41.4 & 6.7 & 22.1 & 68.9 & 50.0 & 80.2 & -\\
    ALIGN & \textbf{94.9} & 76.8 & \textbf{66.1} & 52.1 & \textbf{50.8} & 25.0 & \textbf{41.2} & 74.0 & 55.2 & 83.0 & \textbf{76.4}\\
    OpenCLIP  & 93.5 & 76.2 & 56.4 & 53.7 & 50.3 & 20.8 & 28.8 & 70.9 & 50.5 & 82.3 \\
    CLIP  & \textbf{94.9} & \textbf{77.0} & 56.0 & \textbf{63.0} & 48.3 & \textbf{33.3} & 11.5 &\textbf{79.0} & \textbf{62.3} & \textbf{84.0} & 76.2\\
\midrule
    \multicolumn{8}{l}{\textit{Translated benchmark}} \\
    BriVL  & 72.3 & 35.9 & 18.8 & 25.5 & - & - & - & - & -  & - & 24.3\\
    Wukong & 95.4 & 77.1 & 40.9 & 50.3 & - & - & - & - & -  & - & 55.0\\
    CN-CLIP & \textbf{96.0} & \textbf{79.7} & \textbf{51.2} & \textbf{52.0} & \textbf{55.1} & \textbf{26.2} & \textbf{49.9} & \textbf{79.4} & \textbf{63.5} & \textbf{84.9} & \textbf{59.6} \\
\bottomrule
\end{tabular}
\end{adjustbox}
\caption{Experimental results of the zero-shot image classification performance of models on ICinW. }
\label{tb:zs_v2}
\end{table*}

\subsubsection{Open-Domain Image Classification Benchmark in Chinese}
Contrastive pretraining on image-text pairs builds a connection between vision and natural language. Natural language supervision instead of crowd-sourced labeling endows models with the capability of zero-shot image classification by computing similarities between the given image and the text descriptions of the labels in the candidate set. 
Recent progress in this field is the ELEVATER benchmark~\citep{elevater}. 
The track ICinW for open-domain image classification consists of a series of image classification datasets, including ImageNet~\citep{imagenet}, CIFAR~\citep{cifar}, MNIST~\citep{mnist}, etc.
In order to evaluate Chinese CLIP on the datasets, we first transform the datasets for Chinese models by translating labels and prompts into Chinese. 

\subsubsection{Experimental Results}
Table~\ref{tb:zs_v2} reports the performance of both English models and Chinese models. 
The baselines pretrained on English data include DeCLIP~\citep{declip}, ALIGN~\citep{align}, CLIP~\citep{clip}, and OpenCLIP~\citep{openclip}, and the baselines pretrained on Chinese data include BriVL~\citep{wenlan} and Wukong~\citep{wukong}. 
We report the results of the variant with the best downstream performance for the models. 

We first focus on the comparison with the Chinese baselines. 
To be specific, on all datasets including ImageNet classification, Chinese CLIP surpasses both baselines significantly, and the relative achievements on some datasets are over $100\%$. 
Besides, we also compare Chinese CLIP with the foundation models, e.g., CLIP and ALIGN, which are pretrained on English data. 
It can be found that Chinese CLIP outperforms CLIP or ALIGN on CIFAR-10, CIFAR-100, FER-2013, KITTI-Distance, MNIST, PatchCamelyon, and Pascal-VOC-2007. 
Also, on the classification datasets for general concepts or objects common in both western and eastern culture, Chinese CLIP consistently achieves better performance. 
This indicates that Chinese CLIP is capable of categorizing images to general prototypes. 

However, as to the classification concerned with proper nouns, e.g., FGVC-Aircraft, it is difficult for all models to achieve high accuracy. 
We assume that the related images and texts are not common in the pretraining datasets, and it is also hard for the models to understand the names of airplanes without finetuning. 
Specifically, for Chinese models, the translation or even transliteration can significantly affect the performance of Chinese CLIP. It encourages building a benchmark of ``Image Classification in the Wild for Chinese Models''.

\subsubsection{Analysis}

\paragraph{Sensitivity to Handcrafted Prompts}
While the benchmark ELEVATER provides specific prompts for each dataset, we find that this is not always the best option, in comparison with our baseline, translation of the prompts provided by OpenAI CLIP. 
The baseline with around $90$ prompts performs the best on average. 
However, for some datasets, specific prompts designed with human knowledge can boost the performance significantly. 
A typical case is the classification of airplanes. 
We test $\text{CN-CLIP}\rm_{ViT\text{-}L/14}$ with our specified prompts that are related to the knowledge of aircraft, e.g.,``{label}, a photo of an airplane", ``{label}, a zoomed image of a fighter'', etc., and the translation of the OpenAI prompts.
Experimental results show that the model can achieve an accuracy of $16.0$ with the specified prompts but only $13.8$ with the OpenAI prompts. 

\paragraph{Inability to Understand Negation}
Previous studies~\citep{negbert, bertnot} demonstrate that even strong NLP pretrained models  often make mistakes in negation problems. 
We explore CLIP's capability to understand negation by conducting experiments on KITTI-Distance~\citep{kitti} and PatchCamelyon~\citep{patchcamelyon}. 
KITTI-Distance provides $4$ options for models to judge, including ``next to a car'', ``near a car'', ``at a distance away from a car'', and ``no car''. The last one is concerned with negation. 
We compare the model performance using the text ``no cars'' and ``others'' for the last label. 
We observe that it is hard for the model to understand negation. By changing the label from ``others'' to ``no cars'', the performance drops by $48.1\%$ in accuracy ($49.9$ vs. $25.9$). 
Similarly, in the experiments on PatchCamelyon, the performance drops from $63.5$ to $50.2$ by changing labels from ``mainly red'' and ``green block in the middle'' to ``no green block in the middle'' and ``green block in the middle''. 
This shows the limitation of the training of CLIP in learning negation. 
The texts in the pretraining datasets are mostly descriptions of the images, which indicate their objects or features but often do not indicate the absence of objects. 

\subsection{Deployment}
For deployment, we develop ONNX-based and TensorRT-based models based on our Pytorch-based pretrained Chinese CLIP models. 
As expected, we observe that the inference efficiency increases significantly while there is almost no performance sacrifice. 
Specifically, the inference efficiency of TensorRT-based models is around $2$ to $10$ times faster than the Pytorch-based models. More statistics are listed in Appendix~\ref{appendix:deployment}. 

\section{Related Work}

Previous vision-language pretrained models are mostly BERT/T5-style~\citep{bert, t5}, which involves cross-modal fusion~\citep{uniter, unicoder-vl, visualbert, vilbert, interbert, oscar, pixelbert, e2e-vlp, vinvl, clipvil, simvlm, ofa, albef, blip, mplug, vlmo, beit3}. 
CLIP~\citep{clip}, instead, is a contrastive-learning-based two-tower model, which can serve as a vision foundation model.  
Following CLIP, a series of similar contrastive-learning-based multimodal pretrained models were proposed and reached new SOTAs in cross-modal retrieval and zero-shot classification~\citep{align, filip, florence}. 
Furthermore, CLIP can be adaptive to other models. 
A typical case is that CLIP is essential to many image generation models, e.g., DALL-E~\citep{dalle}, DALL-E 2~\citep{dalle2}, Stable Diffusion~\citep{latent_diffusion}, etc. 
The success of multimodal pretraining encouraged the transfer of the existing methods to Chinese pretraining, including generative pretrained models~\citep{m6, wenlan, m6-t, m6-10t, ofa} and contrastive pretrained models~\citep{wenlan, wukong, r2d2, altclip}. 


\section{Conclusion}
In this work, we propose Chinese CLIP, a Chinese-specific vision-language foundation model. 
Specifically, we construct a pretraining dataset of around $200$ million samples, and pretrain a series of Chinese CLIP models with the proposed two-stage pretraining method, which improves both pretraining efficiency and effectiveness. 
Our comprehensive evaluation shows that Chinese CLIP can reach state-of-the-art performance on multiple cross-modal retrieval datasets in zero-shot learning and finetuning. 
Furthermore, we demonstrate that Chinese CLIP models can also achieve competitive performance in zero-shot image classification across $10$ datasets.

\section*{Limitations}
A number of issues reflect the limitations of this work but also point out some directions for our future research. In this section, we generally discuss some limitations about the scale of data and model. 

\paragraph{Data} 
The core of CLIP pretraining is the simple but effective large-scale contrastive pretraining on extremely large-scale data. Though we have utilized around $200$ million samples, compared with recent studies~\citep{florence, pali} the scale of our pretraining data is relatively small. 
Thus one of our next-step studies is scaling up the quantity of the pretraining data to evaluate the performance improvement with data scaling. 
Furthermore, we still find it hard to decide what a ``high-quality'' dataset for CLIP is. 
In the previous studies~\citep{align, declip}, the preprocessing methods are mostly simple to avoid the loss of data. 
However, there are still many samples where the image and text are not matched properly, which may provide negative information to the pretraining. 
In our future research, we plan to use the pretrained Chinese CLIP model to compute a score for each image-text pair in a larger dataset, filter those whose scores are under the specified threshold, and pretrain the new models with the new data. 
This is one of the possible solutions to explore the relationship between data quality and pretraining effectiveness. 
Also, such cycling might bring continuous performance enhancement in downstream tasks. 

\paragraph{Model} 
Recently we have witnessed that in many domains the scaling of model size can lead to consistent performance improvement~\citep{scaling_laws,emergent}, and in this work, we also find that the scaling of model size for Chinese CLIP can achieve steady performance improvement in different downstream tasks, including retrieval and classification. 
Recent studies have scaled ViT and also CLIP-like models to a much larger scale than our largest $\text{CN-CLIP}\rm_{ViT\text{-}H/14}$, e.g., the 3B Swin-v2~\citep{swin-v2}, the 4B ViT-e~\citep{pali}, etc. 
In the future, we will continue exploring scaling up models in line with scaling up data in order to build a more effective Chinese CLIP. 

Another issue of model scaling connected with the real-world application is how to build effective small models. 
Experimental results show that our smallest Chinese CLIP $\text{CN-CLIP}\rm_{RN50}$ performs much worse than the ViT variants. 
However, in real-world applications, effective small models that are available for deployment are usually more welcomed. 
Thus it is necessary to explore distillation for CLIP so that the capability of large models can be transferred to small models for application.



\section*{Ethics Statement}
The proposed model is a contrastive-learning-based vision-language foundation model, which generates features for images and texts. Those features can be representation of visual and linguistic information, and they can support applications such as search engine, recommender system, etc. 
Besides, this model can play as a foundation model support recent image generation models, e.g., diffusion models~\citep{dalle2}. 
This may create risks as the AI generated contents may reflect harmful information, such as hate, bias, pornography, etc. 
In most cases, these cases should be attributed to the training of the image generation models. 
Still, we cannot avoid the negative effects from the CLIP representations to the generation. 
In the future, we will study how to filter pretraining data to avoid the potential risks. 

\bibliography{anthology,custom}

\begin{thebibliography}{69}
\expandafter\ifx\csname natexlab\endcsname\relax\def\natexlab#1{#1}\fi

\bibitem[{Bommasani et~al.(2021)Bommasani, Hudson, Adeli, Altman, Arora, von
  Arx, Bernstein, Bohg, Bosselut, Brunskill et~al.}]{foundation_model}
Rishi Bommasani, Drew~A Hudson, Ehsan Adeli, Russ Altman, Simran Arora, Sydney
  von Arx, Michael~S Bernstein, Jeannette Bohg, Antoine Bosselut, Emma
  Brunskill, et~al. 2021.
\newblock On the opportunities and risks of foundation models.
\newblock \emph{arXiv preprint arXiv:2108.07258}.

\bibitem[{Bossard et~al.(2014)Bossard, Guillaumin, and Gool}]{food101}
Lukas Bossard, Matthieu Guillaumin, and Luc~Van Gool. 2014.
\newblock Food-101--mining discriminative components with random forests.
\newblock In \emph{European conference on computer vision}, pages 446--461.
  Springer.

\bibitem[{Carlsson et~al.(2022)Carlsson, Eisen, Rekathati, and
  Sahlgren}]{mclip}
Fredrik Carlsson, Philipp Eisen, Faton Rekathati, and Magnus Sahlgren. 2022.
\newblock Cross-lingual and multilingual clip.
\newblock In \emph{Proceedings of the Language Resources and Evaluation
  Conference}, pages 6848--6854, Marseille, France. European Language Resources
  Association.

\bibitem[{Chen et~al.(2022{\natexlab{a}})Chen, Wang, Changpinyo, Piergiovanni,
  Padlewski, Salz, Goodman, Grycner, Mustafa, Beyer et~al.}]{pali}
Xi~Chen, Xiao Wang, Soravit Changpinyo, AJ~Piergiovanni, Piotr Padlewski,
  Daniel Salz, Sebastian Goodman, Adam Grycner, Basil Mustafa, Lucas Beyer,
  et~al. 2022{\natexlab{a}}.
\newblock Pali: A jointly-scaled multilingual language-image model.
\newblock \emph{arXiv preprint arXiv:2209.06794}.

\bibitem[{Chen et~al.(2015)Chen, Fang, Lin, Vedantam, Gupta, Doll{\'{a}}r, and
  Zitnick}]{coco_cap}
Xinlei Chen, Hao Fang, Tsung{-}Yi Lin, Ramakrishna Vedantam, Saurabh Gupta,
  Piotr Doll{\'{a}}r, and C.~Lawrence Zitnick. 2015.
\newblock Microsoft {COCO} captions: Data collection and evaluation server.
\newblock \emph{CoRR}, abs/1504.00325.

\bibitem[{Chen et~al.(2020)Chen, Li, Yu, Kholy, Ahmed, Gan, Cheng, and
  Liu}]{uniter}
Yen{-}Chun Chen, Linjie Li, Licheng Yu, Ahmed~El Kholy, Faisal Ahmed, Zhe Gan,
  Yu~Cheng, and Jingjing Liu. 2020.
\newblock {UNITER:} universal image-text representation learning.
\newblock In \emph{{ECCV} 2020}, volume 12375 of \emph{Lecture Notes in
  Computer Science}, pages 104--120. Springer.

\bibitem[{Chen et~al.(2022{\natexlab{b}})Chen, Liu, Zhang, Ye, Yang, and
  Wu}]{altclip}
Zhong-Yong Chen, Guangyi Liu, Bohan Zhang, Fulong Ye, Qinghong Yang, and
  Ledell~Yu Wu. 2022{\natexlab{b}}.
\newblock Altclip: Altering the language encoder in clip for extended language
  capabilities.
\newblock \emph{ArXiv}, abs/2211.06679.

\bibitem[{Cheng et~al.(2017)Cheng, Han, and Lu}]{resisc45}
Gong Cheng, Junwei Han, and Xiaoqiang Lu. 2017.
\newblock Remote sensing image scene classification: Benchmark and state of the
  art.
\newblock \emph{Proceedings of the IEEE}, 105(10):1865--1883.

\bibitem[{Cimpoi et~al.(2014)Cimpoi, Maji, Kokkinos, Mohamed, and
  Vedaldi}]{dtd}
Mircea Cimpoi, Subhransu Maji, Iasonas Kokkinos, Sammy Mohamed, and Andrea
  Vedaldi. 2014.
\newblock Describing textures in the wild.
\newblock In \emph{Proceedings of the IEEE conference on computer vision and
  pattern recognition}, pages 3606--3613.

\bibitem[{Cubuk et~al.(2019)Cubuk, Zoph, Mane, Vasudevan, and Le}]{autoaugment}
Ekin~D Cubuk, Barret Zoph, Dandelion Mane, Vijay Vasudevan, and Quoc~V Le.
  2019.
\newblock Autoaugment: Learning augmentation strategies from data.
\newblock In \emph{Proceedings of the IEEE/CVF Conference on Computer Vision
  and Pattern Recognition}, pages 113--123.

\bibitem[{Cui et~al.(2020)Cui, Che, Liu, Qin, Wang, and Hu}]{wwm}
Yiming Cui, Wanxiang Che, Ting Liu, Bing Qin, Shijin Wang, and Guoping Hu.
  2020.
\newblock \href {https://www.aclweb.org/anthology/2020.findings-emnlp.58}
  {Revisiting pre-trained models for {C}hinese natural language processing}.
\newblock In \emph{Proceedings of the 2020 Conference on Empirical Methods in
  Natural Language Processing: Findings}, pages 657--668, Online. Association
  for Computational Linguistics.

\bibitem[{Deng et~al.(2009)Deng, Dong, Socher, Li, Li, and
  Fei{-}Fei}]{imagenet}
Jia Deng, Wei Dong, Richard Socher, Li{-}Jia Li, Kai Li, and Li~Fei{-}Fei.
  2009.
\newblock Imagenet: {A} large-scale hierarchical image database.
\newblock In \emph{2009 {IEEE} Computer Society Conference on Computer Vision
  and Pattern Recognition {(CVPR} 2009), 20-25 June 2009, Miami, Florida,
  {USA}}, pages 248--255. {IEEE} Computer Society.

\bibitem[{Deng(2012)}]{mnist}
Li~Deng. 2012.
\newblock The mnist database of handwritten digit images for machine learning
  research [best of the web].
\newblock \emph{IEEE signal processing magazine}, 29(6):141--142.

\bibitem[{Devlin et~al.(2019)Devlin, Chang, Lee, and Toutanova}]{bert}
Jacob Devlin, Ming{-}Wei Chang, Kenton Lee, and Kristina Toutanova. 2019.
\newblock {BERT:} pre-training of deep bidirectional transformers for language
  understanding.
\newblock In \emph{{NAACL-HLT} 2019}, pages 4171--4186. Association for
  Computational Linguistics.

\bibitem[{Everingham et~al.(2010)Everingham, Gool, Williams, Winn, and
  Zisserman}]{voc}
Mark Everingham, Luc~Van Gool, Christopher K.~I. Williams, John~M. Winn, and
  Andrew Zisserman. 2010.
\newblock The pascal visual object classes {(VOC)} challenge.
\newblock \emph{Int. J. Comput. Vis.}, 88(2):303--338.

\bibitem[{Fei et~al.(2021)Fei, Lu, Gao, Yang, Huo, Wen, Lu, Song, Gao, Xiang
  et~al.}]{wenlan}
Nanyi Fei, Zhiwu Lu, Yizhao Gao, Guoxing Yang, Yuqi Huo, Jingyuan Wen, Haoyu
  Lu, Ruihua Song, Xin Gao, Tao Xiang, et~al. 2021.
\newblock Wenlan 2.0: Make ai imagine via a multimodal foundation model.
\newblock \emph{arXiv preprint arXiv:2110.14378}.

\bibitem[{Fei-Fei et~al.(2004)Fei-Fei, Fergus, and Perona}]{caltech}
Li~Fei-Fei, Rob Fergus, and Pietro Perona. 2004.
\newblock Learning generative visual models from few training examples: An
  incremental bayesian approach tested on 101 object categories.
\newblock In \emph{2004 conference on computer vision and pattern recognition
  workshop}, pages 178--178. IEEE.

\bibitem[{Fritsch et~al.(2013)Fritsch, Kuehnl, and Geiger}]{kitti}
Jannik Fritsch, Tobias Kuehnl, and Andreas Geiger. 2013.
\newblock A new performance measure and evaluation benchmark for road detection
  algorithms.
\newblock In \emph{16th International IEEE Conference on Intelligent
  Transportation Systems (ITSC 2013)}, pages 1693--1700. IEEE.

\bibitem[{Gordon et~al.(2021)Gordon, Duh, and Kaplan}]{scaling_laws}
Mitchell~A. Gordon, Kevin Duh, and Jared Kaplan. 2021.
\newblock Data and parameter scaling laws for neural machine translation.
\newblock In \emph{{EMNLP} 2021}, pages 5915--5922. Association for
  Computational Linguistics.

\bibitem[{Gu et~al.(2022)Gu, Meng, Lu, Hou, Niu, Xu, Liang, Zhang, Jiang, and
  Xu}]{wukong}
Jiaxi Gu, Xiaojun Meng, Guansong Lu, Lu~Hou, Minzhe Niu, Hang Xu, Xiaodan
  Liang, Wei Zhang, Xin Jiang, and Chunjing Xu. 2022.
\newblock Wukong: 100 million large-scale chinese cross-modal pre-training
  dataset and a foundation framework.
\newblock \emph{arXiv preprint arXiv:2202.06767}.

\bibitem[{Helber et~al.(2019)Helber, Bischke, Dengel, and Borth}]{eurosat}
Patrick Helber, Benjamin Bischke, Andreas Dengel, and Damian Borth. 2019.
\newblock Eurosat: A novel dataset and deep learning benchmark for land use and
  land cover classification.
\newblock \emph{IEEE Journal of Selected Topics in Applied Earth Observations
  and Remote Sensing}, 12(7):2217--2226.

\bibitem[{Hosseini et~al.(2021)Hosseini, Reddy, Bahdanau, Hjelm, Sordoni, and
  Courville}]{bertnot}
Arian Hosseini, Siva Reddy, Dzmitry Bahdanau, R~Devon Hjelm, Alessandro
  Sordoni, and Aaron Courville. 2021.
\newblock Understanding by understanding not: Modeling negation in language
  models.
\newblock \emph{arXiv preprint arXiv:2105.03519}.

\bibitem[{Huang et~al.(2020)Huang, Zeng, Liu, Fu, and Fu}]{pixelbert}
Zhicheng Huang, Zhaoyang Zeng, Bei Liu, Dongmei Fu, and Jianlong Fu. 2020.
\newblock Pixel-bert: Aligning image pixels with text by deep multi-modal
  transformers.
\newblock \emph{CoRR}, abs/2004.00849.

\bibitem[{Ilharco et~al.(2021)Ilharco, Wortsman, Wightman, Gordon, Carlini,
  Taori, Dave, Shankar, Namkoong, Miller, Hajishirzi, Farhadi, and
  Schmidt}]{openclip}
Gabriel Ilharco, Mitchell Wortsman, Ross Wightman, Cade Gordon, Nicholas
  Carlini, Rohan Taori, Achal Dave, Vaishaal Shankar, Hongseok Namkoong, John
  Miller, Hannaneh Hajishirzi, Ali Farhadi, and Ludwig Schmidt. 2021.
\newblock \href {https://doi.org/10.5281/zenodo.5143773} {Openclip}.
\newblock If you use this software, please cite it as below.

\bibitem[{Jia et~al.(2021)Jia, Yang, Xia, Chen, Parekh, Pham, Le, Sung, Li, and
  Duerig}]{align}
Chao Jia, Yinfei Yang, Ye~Xia, Yi-Ting Chen, Zarana Parekh, Hieu Pham, Quoc~V
  Le, Yunhsuan Sung, Zhen Li, and Tom Duerig. 2021.
\newblock Scaling up visual and vision-language representation learning with
  noisy text supervision.
\newblock \emph{arXiv preprint arXiv:2102.05918}.

\bibitem[{Khandelwal and Sawant(2019)}]{negbert}
Aditya Khandelwal and Suraj Sawant. 2019.
\newblock Negbert: a transfer learning approach for negation detection and
  scope resolution.
\newblock \emph{arXiv preprint arXiv:1911.04211}.

\bibitem[{Kiela et~al.(2020)Kiela, Firooz, Mohan, Goswami, Singh, Ringshia, and
  Testuggine}]{hateful}
Douwe Kiela, Hamed Firooz, Aravind Mohan, Vedanuj Goswami, Amanpreet Singh,
  Pratik Ringshia, and Davide Testuggine. 2020.
\newblock The hateful memes challenge: Detecting hate speech in multimodal
  memes.
\newblock \emph{Advances in Neural Information Processing Systems},
  33:2611--2624.

\bibitem[{Krause et~al.(2013)Krause, Stark, Deng, and Fei-Fei}]{stanford_cars}
Jonathan Krause, Michael Stark, Jia Deng, and Li~Fei-Fei. 2013.
\newblock 3d object representations for fine-grained categorization.
\newblock In \emph{Proceedings of the IEEE international conference on computer
  vision workshops}, pages 554--561.

\bibitem[{Krishna et~al.(2017)Krishna, Zhu, Groth, Johnson, Hata, Kravitz,
  Chen, Kalantidis, Li, Shamma, Bernstein, and Fei{-}Fei}]{vg}
Ranjay Krishna, Yuke Zhu, Oliver Groth, Justin Johnson, Kenji Hata, Joshua
  Kravitz, Stephanie Chen, Yannis Kalantidis, Li{-}Jia Li, David~A. Shamma,
  Michael~S. Bernstein, and Li~Fei{-}Fei. 2017.
\newblock Visual genome: Connecting language and vision using crowdsourced
  dense image annotations.
\newblock \emph{IJCV}, 123(1):32--73.

\bibitem[{Krizhevsky et~al.(2009)Krizhevsky, Hinton et~al.}]{cifar}
Alex Krizhevsky, Geoffrey Hinton, et~al. 2009.
\newblock Learning multiple layers of features from tiny images.

\bibitem[{Lan et~al.(2017)Lan, Li, and Dong}]{flickr30k-cn}
Weiyu Lan, Xirong Li, and Jianfeng Dong. 2017.
\newblock Fluency-guided cross-lingual image captioning.
\newblock \emph{Proceedings of the 25th ACM international conference on
  Multimedia}.

\bibitem[{Li et~al.(2022{\natexlab{a}})Li, Xu, Tian, Wang, Yan, Bi, Ye, Chen,
  Xu, Cao et~al.}]{mplug}
Chenliang Li, Haiyang Xu, Junfeng Tian, Wei Wang, Ming Yan, Bin Bi, Jiabo Ye,
  Hehong Chen, Guohai Xu, Zheng Cao, et~al. 2022{\natexlab{a}}.
\newblock mplug: Effective and efficient vision-language learning by
  cross-modal skip-connections.
\newblock \emph{arXiv preprint arXiv:2205.12005}.

\bibitem[{Li et~al.(2022{\natexlab{b}})Li, Liu, Li, Zhang, Aneja, Yang, Jin,
  Lee, Hu, Liu et~al.}]{elevater}
Chunyuan Li, Haotian Liu, Liunian~Harold Li, Pengchuan Zhang, Jyoti Aneja,
  Jianwei Yang, Ping Jin, Yong~Jae Lee, Houdong Hu, Zicheng Liu, et~al.
  2022{\natexlab{b}}.
\newblock Elevater: A benchmark and toolkit for evaluating language-augmented
  visual models.
\newblock \emph{arXiv preprint arXiv:2204.08790}.

\bibitem[{Li et~al.(2019{\natexlab{a}})Li, Duan, Fang, Jiang, and
  Zhou}]{unicoder-vl}
Gen Li, Nan Duan, Yuejian Fang, Daxin Jiang, and Ming Zhou. 2019{\natexlab{a}}.
\newblock Unicoder-vl: {A} universal encoder for vision and language by
  cross-modal pre-training.
\newblock \emph{CoRR}, abs/1908.06066.

\bibitem[{Li et~al.(2022{\natexlab{c}})Li, Li, Xiong, and Hoi}]{blip}
Junnan Li, Dongxu Li, Caiming Xiong, and Steven Hoi. 2022{\natexlab{c}}.
\newblock Blip: Bootstrapping language-image pre-training for unified
  vision-language understanding and generation.
\newblock \emph{arXiv preprint arXiv:2201.12086}.

\bibitem[{Li et~al.(2021{\natexlab{a}})Li, Selvaraju, Gotmare, Joty, Xiong, and
  Hoi}]{albef}
Junnan Li, Ramprasaath~R. Selvaraju, Akhilesh Gotmare, Shafiq~R. Joty, Caiming
  Xiong, and Steven~Chu{-}Hong Hoi. 2021{\natexlab{a}}.
\newblock Align before fuse: Vision and language representation learning with
  momentum distillation.
\newblock In \emph{NeurIPS 2021}, pages 9694--9705.

\bibitem[{Li et~al.(2019{\natexlab{b}})Li, Yatskar, Yin, Hsieh, and
  Chang}]{visualbert}
Liunian~Harold Li, Mark Yatskar, Da~Yin, Cho-Jui Hsieh, and Kai-Wei Chang.
  2019{\natexlab{b}}.
\newblock Visualbert: A simple and performant baseline for vision and language.
\newblock \emph{ArXiv}, abs/1908.03557.

\bibitem[{Li et~al.(2019{\natexlab{c}})Li, Xu, Wang, Lan, Jia, Yang, and
  Xu}]{coco-cn}
Xirong Li, Chaoxi Xu, Xiaoxu Wang, Weiyu Lan, Zhengxiong Jia, Gang Yang, and
  Jieping Xu. 2019{\natexlab{c}}.
\newblock Coco-cn for cross-lingual image tagging, captioning, and retrieval.
\newblock \emph{IEEE Transactions on Multimedia}, 21:2347--2360.

\bibitem[{Li et~al.(2020)Li, Yin, Li, Hu, Zhang, Zhang, Wang, Hu, Dong, Wei,
  Choi, and Gao}]{oscar}
Xiujun Li, Xi~Yin, Chunyuan Li, Xiaowei Hu, Pengchuan Zhang, Lei Zhang, Lijuan
  Wang, Houdong Hu, Li~Dong, Furu Wei, Yejin Choi, and Jianfeng Gao. 2020.
\newblock Oscar: Object-semantics aligned pre-training for vision-language
  tasks.
\newblock In \emph{ECCV}.

\bibitem[{Li et~al.(2021{\natexlab{b}})Li, Liang, Zhao, Cui, Ouyang, Shao, Yu,
  and Yan}]{declip}
Yangguang Li, Feng Liang, Lichen Zhao, Yufeng Cui, Wanli Ouyang, Jing Shao,
  Fengwei Yu, and Junjie Yan. 2021{\natexlab{b}}.
\newblock Supervision exists everywhere: A data efficient contrastive
  language-image pre-training paradigm.
\newblock \emph{arXiv preprint arXiv:2110.05208}.

\bibitem[{Lin et~al.(2021{\natexlab{a}})Lin, Men, Yang, Zhou, Ding, Zhang,
  Wang, Wang, Jiang, Jia, Zhang, Zhang, Zou, Li, Deng, Liu, Xue, Zhou, Ma, Yu,
  Li, Lin, Zhou, Tang, and Yang}]{m6}
Junyang Lin, Rui Men, An~Yang, Chang Zhou, Ming Ding, Yichang Zhang, Peng Wang,
  Ang Wang, Le~Jiang, Xianyan Jia, Jie Zhang, Jianwei Zhang, Xu~Zou, Zhikang
  Li, Xiaodong Deng, Jie Liu, Jinbao Xue, Huiling Zhou, Jianxin Ma, Jin Yu,
  Yong Li, Wei Lin, Jingren Zhou, Jie Tang, and Hongxia Yang.
  2021{\natexlab{a}}.
\newblock {M6:} {A} chinese multimodal pretrainer.
\newblock \emph{CoRR}, abs/2103.00823.

\bibitem[{Lin et~al.(2021{\natexlab{b}})Lin, Yang, Bai, Zhou, Jiang, Jia, Wang,
  Zhang, Li, Lin, Zhou, and Yang}]{m6-10t}
Junyang Lin, An~Yang, Jinze Bai, Chang Zhou, Le~Jiang, Xianyan Jia, Ang Wang,
  Jie Zhang, Yong Li, Wei Lin, Jingren Zhou, and Hongxia Yang.
  2021{\natexlab{b}}.
\newblock {M6-10T:} {A} sharing-delinking paradigm for efficient multi-trillion
  parameter pretraining.
\newblock \emph{CoRR}, abs/2110.03888.

\bibitem[{Lin et~al.(2020)Lin, Yang, Zhang, Liu, Zhou, and Yang}]{interbert}
Junyang Lin, An~Yang, Yichang Zhang, Jie Liu, Jingren Zhou, and Hongxia Yang.
  2020.
\newblock Interbert: Vision-and-language interaction for multi-modal
  pretraining.
\newblock \emph{CoRR}, abs/2003.13198.

\bibitem[{Liu et~al.(2022)Liu, Hu, Lin, Yao, Xie, Wei, Ning, Cao, Zhang, Dong
  et~al.}]{swin-v2}
Ze~Liu, Han Hu, Yutong Lin, Zhuliang Yao, Zhenda Xie, Yixuan Wei, Jia Ning, Yue
  Cao, Zheng Zhang, Li~Dong, et~al. 2022.
\newblock Swin transformer v2: Scaling up capacity and resolution.
\newblock In \emph{Proceedings of the IEEE/CVF Conference on Computer Vision
  and Pattern Recognition}, pages 12009--12019.

\bibitem[{Lu et~al.(2019)Lu, Batra, Parikh, and Lee}]{vilbert}
Jiasen Lu, Dhruv Batra, Devi Parikh, and Stefan Lee. 2019.
\newblock Vilbert: Pretraining task-agnostic visiolinguistic representations
  for vision-and-language tasks.
\newblock In \emph{NeurIPS 2019}, pages 13--23.

\bibitem[{Maji et~al.(2013)Maji, Rahtu, Kannala, Blaschko, and Vedaldi}]{fgvc}
Subhransu Maji, Esa Rahtu, Juho Kannala, Matthew Blaschko, and Andrea Vedaldi.
  2013.
\newblock Fine-grained visual classification of aircraft.
\newblock \emph{arXiv preprint arXiv:1306.5151}.

\bibitem[{Nilsback and Zisserman(2008)}]{flowers}
Maria-Elena Nilsback and Andrew Zisserman. 2008.
\newblock Automated flower classification over a large number of classes.
\newblock In \emph{2008 Sixth Indian Conference on Computer Vision, Graphics \&
  Image Processing}, pages 722--729. IEEE.

\bibitem[{Parkhi et~al.(2012)Parkhi, Vedaldi, Zisserman, and Jawahar}]{pets}
Omkar~M. Parkhi, Andrea Vedaldi, Andrew Zisserman, and C.~V. Jawahar. 2012.
\newblock Cats and dogs.
\newblock In \emph{2012 {IEEE} Conference on Computer Vision and Pattern
  Recognition}, pages 3498--3505. {IEEE} Computer Society.

\bibitem[{Radford et~al.(2021)Radford, Kim, Hallacy, Ramesh, Goh, Agarwal,
  Sastry, Askell, Mishkin, Clark, Krueger, and Sutskever}]{clip}
Alec Radford, Jong~Wook Kim, Chris Hallacy, Aditya Ramesh, Gabriel Goh,
  Sandhini Agarwal, Girish Sastry, Amanda Askell, Pamela Mishkin, Jack Clark,
  Gretchen Krueger, and Ilya Sutskever. 2021.
\newblock Learning transferable visual models from natural language
  supervision.
\newblock In \emph{{ICML} 2021}, volume 139 of \emph{Proceedings of Machine
  Learning Research}, pages 8748--8763. {PMLR}.

\bibitem[{Raffel et~al.(2020)Raffel, Shazeer, Roberts, Lee, Narang, Matena,
  Zhou, Li, and Liu}]{t5}
Colin Raffel, Noam Shazeer, Adam Roberts, Katherine Lee, Sharan Narang, Michael
  Matena, Yanqi Zhou, Wei Li, and Peter~J. Liu. 2020.
\newblock Exploring the limits of transfer learning with a unified text-to-text
  transformer.
\newblock \emph{J. Mach. Learn. Res.}, 21:140:1--140:67.

\bibitem[{Ramesh et~al.(2022)Ramesh, Dhariwal, Nichol, Chu, and Chen}]{dalle2}
Aditya Ramesh, Prafulla Dhariwal, Alex Nichol, Casey Chu, and Mark Chen. 2022.
\newblock Hierarchical text-conditional image generation with clip latents.
\newblock \emph{arXiv preprint arXiv:2204.06125}.

\bibitem[{Ramesh et~al.(2021)Ramesh, Pavlov, Goh, Gray, Voss, Radford, Chen,
  and Sutskever}]{dalle}
Aditya Ramesh, Mikhail Pavlov, Gabriel Goh, Scott Gray, Chelsea Voss, Alec
  Radford, Mark Chen, and Ilya Sutskever. 2021.
\newblock Zero-shot text-to-image generation.
\newblock In \emph{{ICML} 2021}, volume 139 of \emph{Proceedings of Machine
  Learning Research}, pages 8821--8831. {PMLR}.

\bibitem[{Rombach et~al.(2022)Rombach, Blattmann, Lorenz, Esser, and
  Ommer}]{latent_diffusion}
Robin Rombach, Andreas Blattmann, Dominik Lorenz, Patrick Esser, and Bj{\"o}rn
  Ommer. 2022.
\newblock High-resolution image synthesis with latent diffusion models.
\newblock In \emph{Proceedings of the IEEE/CVF Conference on Computer Vision
  and Pattern Recognition}, pages 10684--10695.

\bibitem[{Schuhmann et~al.(2021)Schuhmann, Vencu, Beaumont, Kaczmarczyk,
  Mullis, Katta, Coombes, Jitsev, and Komatsuzaki}]{schuhmann2021laion}
Christoph Schuhmann, Richard Vencu, Romain Beaumont, Robert Kaczmarczyk,
  Clayton Mullis, Aarush Katta, Theo Coombes, Jenia Jitsev, and Aran
  Komatsuzaki. 2021.
\newblock Laion-400m: Open dataset of clip-filtered 400 million image-text
  pairs.
\newblock \emph{arXiv preprint arXiv:2111.02114}.

\bibitem[{Shen et~al.(2021)Shen, Li, Tan, Bansal, Rohrbach, Chang, Yao, and
  Keutzer}]{clipvil}
Sheng Shen, Liunian~Harold Li, Hao Tan, Mohit Bansal, Anna Rohrbach, Kai-Wei
  Chang, Zhewei Yao, and Kurt Keutzer. 2021.
\newblock How much can clip benefit vision-and-language tasks?
\newblock \emph{arXiv preprint arXiv:2107.06383}.

\bibitem[{Stallkamp et~al.(2011)Stallkamp, Schlipsing, Salmen, and
  Igel}]{gtsrb}
Johannes Stallkamp, Marc Schlipsing, Jan Salmen, and Christian Igel. 2011.
\newblock The german traffic sign recognition benchmark: {A} multi-class
  classification competition.
\newblock In \emph{{IJCNN} 2011}, pages 1453--1460. {IEEE}.

\bibitem[{Veeling et~al.(2018)Veeling, Linmans, Winkens, Cohen, and
  Welling}]{patchcamelyon}
Bastiaan~S Veeling, Jasper Linmans, Jim Winkens, Taco Cohen, and Max Welling.
  2018.
\newblock Rotation equivariant cnns for digital pathology.
\newblock In \emph{International Conference on Medical image computing and
  computer-assisted intervention}, pages 210--218. Springer.

\bibitem[{Wang et~al.(2022{\natexlab{a}})Wang, Yang, Men, Lin, Bai, Li, Ma,
  Zhou, Zhou, and Yang}]{ofa}
Peng Wang, An~Yang, Rui Men, Junyang Lin, Shuai Bai, Zhikang Li, Jianxin Ma,
  Chang Zhou, Jingren Zhou, and Hongxia Yang. 2022{\natexlab{a}}.
\newblock Unifying architectures, tasks, and modalities through a simple
  sequence-to-sequence learning framework.
\newblock \emph{CoRR}, abs/2202.03052.

\bibitem[{Wang et~al.(2022{\natexlab{b}})Wang, Bao, Dong, Bjorck, Peng, Liu,
  Aggarwal, Mohammed, Singhal, Som et~al.}]{beit3}
Wenhui Wang, Hangbo Bao, Li~Dong, Johan Bjorck, Zhiliang Peng, Qiang Liu, Kriti
  Aggarwal, Owais~Khan Mohammed, Saksham Singhal, Subhojit Som, et~al.
  2022{\natexlab{b}}.
\newblock Image as a foreign language: Beit pretraining for all vision and
  vision-language tasks.
\newblock \emph{arXiv preprint arXiv:2208.10442}.

\bibitem[{Wang et~al.(2021{\natexlab{a}})Wang, Bao, Dong, and Wei}]{vlmo}
Wenhui Wang, Hangbo Bao, Li~Dong, and Furu Wei. 2021{\natexlab{a}}.
\newblock Vlmo: Unified vision-language pre-training with
  mixture-of-modality-experts.
\newblock \emph{CoRR}, abs/2111.02358.

\bibitem[{Wang et~al.(2021{\natexlab{b}})Wang, Yu, Yu, Dai, Tsvetkov, and
  Cao}]{simvlm}
Zirui Wang, Jiahui Yu, Adams~Wei Yu, Zihang Dai, Yulia Tsvetkov, and Yuan Cao.
  2021{\natexlab{b}}.
\newblock Simvlm: Simple visual language model pretraining with weak
  supervision.
\newblock \emph{CoRR}, abs/2108.10904.

\bibitem[{Wei et~al.(2022)Wei, Tay, Bommasani, Raffel, Zoph, Borgeaud,
  Yogatama, Bosma, Zhou, Metzler et~al.}]{emergent}
Jason Wei, Yi~Tay, Rishi Bommasani, Colin Raffel, Barret Zoph, Sebastian
  Borgeaud, Dani Yogatama, Maarten Bosma, Denny Zhou, Donald Metzler, et~al.
  2022.
\newblock Emergent abilities of large language models.
\newblock \emph{arXiv preprint arXiv:2206.07682}.

\bibitem[{Xie et~al.(2022)Xie, Cai, Song, Li, Kong, Wu, Morimitsu, Yao, Wang,
  Leng et~al.}]{r2d2}
Chunyu Xie, Heng Cai, Jianfei Song, Jincheng Li, Fanjing Kong, Xiaoyu Wu,
  Henrique Morimitsu, Lin Yao, Dexin Wang, Dawei Leng, et~al. 2022.
\newblock Zero and r2d2: A large-scale chinese cross-modal benchmark and a
  vision-language framework.
\newblock \emph{arXiv preprint arXiv:2205.03860}.

\bibitem[{Xu et~al.(2021)Xu, Yan, Li, Bi, Huang, Xiao, and Huang}]{e2e-vlp}
Haiyang Xu, Ming Yan, Chenliang Li, Bin Bi, Songfang Huang, Wenming Xiao, and
  Fei Huang. 2021.
\newblock E2e-vlp: End-to-end vision-language pre-training enhanced by visual
  learning.
\newblock \emph{arXiv preprint arXiv:2106.01804}.

\bibitem[{Yang et~al.(2021)Yang, Lin, Men, Zhou, Jiang, Jia, Wang, Zhang, Wang,
  Li, Zhang, Lin, Qu, Zhou, and Yang}]{m6-t}
An~Yang, Junyang Lin, Rui Men, Chang Zhou, Le~Jiang, Xianyan Jia, Ang Wang, Jie
  Zhang, Jiamang Wang, Yong Li, Di~Zhang, Wei Lin, Lin Qu, Jingren Zhou, and
  Hongxia Yang. 2021.
\newblock Exploring sparse expert models and beyond.
\newblock \emph{CoRR}, abs/2105.15082.

\bibitem[{Yao et~al.(2021)Yao, Huang, Hou, Lu, Niu, Xu, Liang, Li, Jiang, and
  Xu}]{filip}
Lewei Yao, Runhui Huang, Lu~Hou, Guansong Lu, Minzhe Niu, Hang Xu, Xiaodan
  Liang, Zhenguo Li, Xin Jiang, and Chunjing Xu. 2021.
\newblock {FILIP:} fine-grained interactive language-image pre-training.
\newblock \emph{CoRR}, abs/2111.07783.

\bibitem[{Yuan et~al.(2021)Yuan, Chen, Chen, Codella, Dai, Gao, Hu, Huang, Li,
  Li, Liu, Liu, Liu, Lu, Shi, Wang, Wang, Xiao, Xiao, Yang, Zeng, Zhou, and
  Zhang}]{florence}
Lu~Yuan, Dongdong Chen, Yi{-}Ling Chen, Noel Codella, Xiyang Dai, Jianfeng Gao,
  Houdong Hu, Xuedong Huang, Boxin Li, Chunyuan Li, Ce~Liu, Mengchen Liu,
  Zicheng Liu, Yumao Lu, Yu~Shi, Lijuan Wang, Jianfeng Wang, Bin Xiao, Zhen
  Xiao, Jianwei Yang, Michael Zeng, Luowei Zhou, and Pengchuan Zhang. 2021.
\newblock Florence: {A} new foundation model for computer vision.
\newblock \emph{CoRR}, abs/2111.11432.

\bibitem[{Zhai et~al.(2022)Zhai, Wang, Mustafa, Steiner, Keysers, Kolesnikov,
  and Beyer}]{lit}
Xiaohua Zhai, Xiao Wang, Basil Mustafa, Andreas Steiner, Daniel Keysers,
  Alexander Kolesnikov, and Lucas Beyer. 2022.
\newblock Lit: Zero-shot transfer with locked-image text tuning.
\newblock In \emph{Proceedings of the IEEE/CVF Conference on Computer Vision
  and Pattern Recognition}, pages 18123--18133.

\bibitem[{Zhang et~al.(2021)Zhang, Li, Hu, Yang, Zhang, Wang, Choi, and
  Gao}]{vinvl}
Pengchuan Zhang, Xiujun Li, Xiaowei Hu, Jianwei Yang, Lei Zhang, Lijuan Wang,
  Yejin Choi, and Jianfeng Gao. 2021.
\newblock Vinvl: Revisiting visual representations in vision-language models.
\newblock In \emph{{CVPR} 2021}, pages 5579--5588. Computer Vision Foundation /
  {IEEE}.

\end{thebibliography}
\bibliographystyle{acl_natbib}

\newpage
\appendix

\section{Appendix}
\label{sec:implementation_details}

\subsection{Model Architecture Details}
\label{subsec:model_hyperparams}

We develop $5$ Chinese CLIP models of different sizes, spanning from around $77$ to $958$ million parameters. 
We include $1$ ResNet-50 model $\text{CN-CLIP}\rm_{RN50}$ and $4$ ViT models, i.e., $\text{CN-CLIP}\rm_{ViT\text{-}B/16}$, $\text{CN-CLIP}\rm_{ViT\text{-}L/14}$, $\text{CN-CLIP}\rm_{ViT\text{-}L/14@336px}$ and $\text{CN-CLIP}\rm_{ViT\text{-}H/14}$, where models are pretrained on images of the resolution of $224 \times 224$ without specification. 
Table~\ref{tb:modelcard} presents the details of the model architecture. 
The smallest model $\text{CN-CLIP}\rm_{RN50}$ consists of a ResNet-50 for the image encoder and a RBT3 for the text encoder. 
The base-size model $\text{CN-CLIP}\rm_{ViT\text{-}B/16}$ consists of a ViT-B/16@224px for the image encoder and a RoBERTa-wwm-Base for the text encoder. 
The large-size model $\text{CN-CLIP}\rm_{ViT\text{-}L/14}$ consists of a ViT-L/14@224px for the image encoder and a RoBERTa-wwm-Base for the text encoder, while $\text{CN-CLIP}\rm_{ViT\text{-}L/14@336px}$ consists of a ViT-L/14@336px and a RoBERTa-wwm-Base. 
Specifically, we pretrain $\text{CN-CLIP}\rm_{ViT\text{-}L/14@336px}$ by continuing pretraining on the pretrained $\text{CN-CLIP}\rm_{ViT\text{-}L/14}$. 
For the adaptation to a larger resolution, we initialize the image positional embedding by applying interpolation to the positional embedding of $\text{CN-CLIP}\rm_{ViT\text{-}L/14}$, following \citet{openclip}. 
The huge-size model $\text{CN-CLIP}\rm_{ViT\text{-}H/14}$ consists of a ViT-H/14 for the image encoder and RoBERTa-wwm-Large for the text encoder. 
More implementation details are presented in Appendix~\ref{sec:implementation_details}.

\begin{table*}[t]
\center
\small
\vskip 0.15in
\begin{adjustbox}{max width=1.\textwidth}
\begin{tabular}{@{\extracolsep{\fill}}lcccccc}
\toprule
  Model
  & \#Params (All)
  & Backbone (I)
  & \#Params (I)
  & Backbone (T)
  & \#Params (T)
  & Resolution
  \\
\midrule
    $\text{CN-CLIP}\rm_{RN50}$
    & 77M & ResNet-50 & 38M & RBT3 & 39M & $224\times224$
    \\
    $\text{CN-CLIP}\rm_{ViT\text{-}B/16}$
    & 188M & ViT-B/16 & 86M & RoBERTa-wwm-Base & 102M & $224\times224$
    \\
    $\text{CN-CLIP}\rm_{ViT\text{-}L/14}$
    & 406M & ViT-L/14 & 304M & RoBERTa-wwm-Base & 102M & $224\times224$
    \\
    $\text{CN-CLIP}\rm_{ViT\text{-}L/14@336px}$
    & 407M & ViT-L/14 & 304M & RoBERTa-wwm-Base & 102M & $336\times336$
    \\
    $\text{CN-CLIP}\rm_{ViT\text{-}H/14}$
    & 958M & ViT-H/14 & 632M & RoBERTa-wwm-Large & 326M & $224\times224$
    \\
\bottomrule
\end{tabular}
\end{adjustbox}
\caption{Hyperparameters of Chinese CLIP models of different sizes. }
\label{tb:modelcard}
\end{table*}

\begin{table*}[t]
\center
\small
\begin{tabular}{@{\extracolsep{\fill}}lccccccc}
\toprule
  \multirow{2}{*}{Model}
  & Embedding
  & \multicolumn{3}{c}{Vision Transformer}
  & \multicolumn{3}{c}{Text Transformer}
  \\
  & dimension
  & layers
  & width
  & heads
  & layers
  & width
  & heads
  \\
\midrule
    $\text{CN-CLIP}\rm_{ViT\text{-}B/16}$
    & 512
    & 12
    & 768
    & 12
    & 12
    & 768
    & 12
    \\
    $\text{CN-CLIP}\rm_{ViT\text{-}L/14}$
    & 768
    & 24
    & 1,024
    & 16
    & 12
    & 768
    & 12
    \\
    $\text{CN-CLIP}\rm_{ViT\text{-}L/14@336px}$
    & 768
    & 24
    & 1,024
    & 16
    & 12
    & 768
    & 12
    \\
    $\text{CN-CLIP}\rm_{ViT\text{-}H/14}$
    & 1,024
    & 32
    & 1,280
    & 16
    & 24
    & 1,024
    & 24
    \\
\bottomrule
\end{tabular}
\caption{Detailed architecture hyperparameters of ViT-based CN-CLIP models.}
\label{tb:detailed_architecture_vit}
\end{table*}

\begin{table*}[t]
\center
\small
\begin{tabular}{@{\extracolsep{\fill}}lcccccc}
\toprule
  \multirow{2}{*}{Model}
  & Embedding
  & \multicolumn{2}{c}{ResNet}
  & \multicolumn{3}{c}{Text Transformer}
  \\
  & dimension
  & blocks
  & width
  & layers
  & width
  & heads
  \\
\midrule
    $\text{CN-CLIP}\rm_{RN50}$
    & 1,024
    & (3, 4, 6, 3)
    & 2,048
    & 3
    & 768
    & 12
    \\
\bottomrule
\end{tabular}
\caption{Detailed architecture hyperparameters of ResNet-based $\text{CN-CLIP}\rm_{RN50}$.}
\label{tb:detailed_architecture_rn}
\end{table*}
We provide more details of their model architectures in Table~\ref{tb:detailed_architecture_vit} and Table~\ref{tb:detailed_architecture_rn}. We keep the architecture of ResNet-50, ViT-B/16, and ViT-L/14 backbones conformed with OpenAI CLIP and the architecture of ViT-H/14 same with LAION CLIP\footnote{\url{https://laion.ai/blog/large-openclip/}}. 
This enables us to initialize the Chinese CLIP image encoders with their weights. 
The text encoders are Chinese Roberta models~\cite{wwm}. Specifically, our most lightweight tiny-size Chinese CLIP uses the architecture of the $3$-layer RBT3 model. The base-size and large-size Chinese CLIP models use the $12$-layer architecture of RoBERTa-wwm-Base. For the huge-size CN-CLIP, we use the $24$-layer architecture of RoBERTa-wwm-Large. The vocabulary size of text tokenizer is $21,128$.

\subsection{Pretraining Details}
\label{subsec:pretraining_details}

\paragraph{Initialization} As mentioned in Section~\ref{subsec:pretraining_method}, we initialize the image encoders of $\text{CN-CLIP}\rm_{RN50}$, $\text{CN-CLIP}\rm_{ViT\text{-}B/16}$ and $\text{CN-CLIP}\rm_{ViT\text{-}L/14}$ using the OpenAI CLIP weights. The image encoder of $\text{CN-CLIP}\rm_{ViT\text{-}H/14}$ is initialized with LAION CLIP. 
Besides the ResNet or ViT parameters, the temperature and visual output projection parameters are also initialized with the pretrained CLIP weights. 
For the text encoder, we initialize the parameters using the released Chinese Roberta weights of the corresponding model scale, with their pooler weights discarded. The text output projection weight is randomly initialized with normal distribution.

\begin{table}[t]
\center
\small
\begin{tabular}{lc}
\toprule
  Hyperparameters
  & Value
  \\
\midrule
    Batch size
    & $32,768$
    \\
    Maximum text length
    & $50$
    \\             
    Peak learning rate
    & $1e-4$
    \\
    Learning rate schedule
    & Cosine
    \\    
    Maximum temperature
    & $100$
    \\
    Weight decay
    & $1e-3$
    \\        
    Warmup iterations
    & $5,000$
    \\        
    Adam $\beta_1$
    & $0.9$
    \\      
    Adam $\beta_2$
    & $0.999$ (ResNet), $0.98$ (ViT)
    \\     
    Adam $\epsilon$
    & $1e-8$ (ResNet), $1e-6$ (ViT)
    \\        
\bottomrule
\end{tabular}
\caption{Common pretraining hyperparameters in the first stage.}
\label{tb:hyperparams_stage1}
\end{table}

\paragraph{Stage 1} 
The pretraining hyperparameters of Stage 1 are shown in Table~\ref{tb:hyperparams_stage1}, which are shared for $\text{CN-CLIP}\rm_{RN50}$, $\text{CN-CLIP}\rm_{ViT\text{-}B/16}$, $\text{CN-CLIP}\rm_{ViT\text{-}L/14}$ and $\text{CN-CLIP}\rm_{ViT\text{-}H/14}$. 
The values of hyperparameters are generally similar to those in OpenAI CLIP~\citep{clip}. 
As to data augmentation, we use random resize cropping and AutoAugment~\citep{autoaugment} on input images. We leverage all-gather communications across GPU workers to compute contrastive loss on the global batch. The above $4$ models are pretrained for around $20$, $44$, $64$, and $26$ epochs in this stage respectively, with the image encoder frozen. The running variance and mean of batch normalization layers are not updated in this stage for $\text{CN-CLIP}\rm_{RN50}$. The optimal epochs of pretraining are determined by measuring the mean-recall under the $3$ downstream zero-shot retrieval tasks during training. Mixed-precision training is activated. 
In this stage, we pretrain $1.6$ days using $64$ NVIDIA V100 GPUs for $\text{CN-CLIP}\rm_{RN50}$, $4.5$ days using $128$ NVIDIA V100 GPUs for $\text{CN-CLIP}\rm_{ViT\text{-}B/16}$, $11.5$ days using $128$ NVIDIA V100 GPUs for $\text{CN-CLIP}\rm_{ViT\text{-}L/14}$ and $3.8$ days using $184$ NVIDIA A100 GPUs for $\text{CN-CLIP}\rm_{ViT\text{-}H/14}$.

\paragraph{Stage 2} 
In Stage 2, we unfreeze the image encoder and update all the model parameters. Except for the peak learning rate, batch size and training epochs, all other hyperparameters mentioned in Stage 1 are kept unchanged. We decrease the learning rate to $2e-5$ for subtler optimization. 
For $\text{CN-CLIP}\rm_{RN50}$, $\text{CN-CLIP}\rm_{ViT\text{-}B/16}$ and $\text{CN-CLIP}\rm_{ViT\text{-}L/14}$, the batch size is shrunk to $16,384$, $16,384$ and $4,608$ respectively due to the limitation in GPU memory. When scaling to $\text{CN-CLIP}\rm_{ViT\text{-}H/14}$, we implement gradient checkpointing, which enables a larger batch size of $32,768$. These $4$ models are pretrained for around $44$, $15$, $7$ and $7$ epochs in Stage 2, respectively.
In this stage, we pretrain $\text{CN-CLIP}\rm_{RN50}$ for $5.8$ days using $64$ NVIDIA V100 GPUs, $\text{CN-CLIP}\rm_{ViT\text{-}B/16}$ for $3.0$ days using $128$ NVIDIA V100 GPUs, $\text{CN-CLIP}\rm_{ViT\text{-}L/14}$ for $8.0$ days using $128$ Nvidia V100 GPUs, and $\text{CN-CLIP}\rm_{ViT\text{-}H/14}$ for $2.2$ days using $184$ NVIDIA A100 GPUs. 

To pretrain a model of a larger resolution, we implement interpolation to the image positional embedding of $\text{CN-CLIP}\rm_{ViT\text{-}L/14}$ for adapting to a larger resolution and continue pretraining with images of the resolution of $336 \times 336$. 
We start from $\text{CN-CLIP}\rm_{ViT\text{-}L/14}$ and continue pretraining by $2$ epochs. 
The pretraining only costs the use of $128$ NVIDIA A100 GPUs for $0.7$ days.

\subsection{Finetuning Details}

\begin{table*}[t]
\center
\begin{adjustbox}{max width=1.\textwidth}
\begin{tabular}{@{\extracolsep{\fill}}lccccccccccccc}
\toprule
  \multirow{2}{*}{Model}
  & \multicolumn{3}{c}{Batch size}
  & \multicolumn{3}{c}{Peak learning rate}
  & \multicolumn{3}{c}{Maximum epochs}
  & \multicolumn{3}{c}{Warmup iterations}

  \\
  & MUGE
  & Flickr
  & COCO
  & MUGE
  & Flickr
  & COCO
  & MUGE
  & Flickr
  & COCO
  & MUGE
  & Flickr
  & COCO
  \\
\midrule
    $\text{CN-CLIP}\rm_{RN50}$
    & 24,576
    & 24,576
    & 24,576
    & 5e-5
    & 6e-5
    & 5e-5
    & 60
    & 30
    & 40
    & 100
    & 20
    & 6
    \\
    $\text{CN-CLIP}\rm_{ViT\text{-}B/16}$
    & 12,800
    & 7,680
    & 12,800
    & 2e-5
    & 5e-5
    & 5e-5
    & 20
    & 16
    & 30
    & 40
    & 20
    & 6
    \\
    $\text{CN-CLIP}\rm_{ViT\text{-}L/14}$
    & 4,096
    & 4,096
    & 4,096
    & 3e-5
    & 2e-5
    & 6e-5
    & 20
    & 16
    & 18
    & 100
    & 60
    & 9
    \\
    $\text{CN-CLIP}\rm_{ViT\text{-}L/14@336px}$
    & 8,192
    & 8,192
    & 8,192
    & 2e-5
    & 2e-5
    & 4e-5
    & 20
    & 18
    & 18
    & 100
    & 20
    & 2
    \\
    $\text{CN-CLIP}\rm_{ViT\text{-}H/14}$
    & 20,480
    & 4,096
    & 5,120
    & 2e-5
    & 6e-6
    & 2e-5
    & 20
    & 18
    & 18
    & 20
    & 6
    & 10
    \\
\bottomrule
\end{tabular}
\end{adjustbox}
\caption{Detailed finetuning hyperparameters of CN-CLIP models.}
\label{tb:detailed_finetune}
\end{table*}

\begin{table*}[t]
\center
\small
\begin{adjustbox}{max width=1.\textwidth}
\begin{tabular}{@{\extracolsep{\fill}}lcccccccc}
\toprule
  Tasks
  &\multicolumn{3}{c}{Text-to-Image}
  &\multicolumn{3}{c}{Image-to-Text}
  & 
  \\
\midrule
  Metrics & R@1 & R@5 & R@10 & R@1 & R@5 & R@10 & MR
  \\
\midrule
  $\text{R2D2}\rm_{ViT\text{-}B}$
  & 42.2    & 69.4  & 77.8	& 43.4	& 69.8	& 78.4 & 63.5
  \\
  $\text{R2D2}\rm_{ViT\text{-}L/14}$
  & 60.7    & 82.0  & 86.9  & 61.5  & 82.9  & 87.7  & 77.0
  \\
  $\text{CN-CLIP}\rm_{ViT\text{-}B/16}$
  & 55.4    & 79.0  & 85.2  & 56.6  & 79.5  & 85.6  & 73.5
  \\
  $\text{CN-CLIP}\rm_{ViT\text{-}L/14}$
  & \textbf{61.6}    & \textbf{83.6}  & \textbf{89.0}  & \textbf{62.5}  & \textbf{83.9}  & \textbf{89.1}  & \textbf{78.3}
  \\

\bottomrule
\end{tabular}
\end{adjustbox}
\caption{Finetuning results on ICR dataset. We report the performance of baselines, $\text{CN-CLIP}\rm_{ViT\text{-}B/16}$ and $\text{CN-CLIP}\rm_{ViT\text{-}L/14}$ on text-to-image and image-to-text retrieval. }
\label{tb:ICR}
\end{table*}
As reported in Table~\ref{tb:muge}, \ref{tb:flickr} and \ref{tb:coco}, we mainly finetune CN-CLIP on $3$ cross-modal retrieval datasets: MUGE, Flickr30K-CN, and COCO-CN.
Most finetuning experiments are conducted on $32$ NVIDIA A100 GPUs.
The finetuning strategy and loss are consistent with the pretraining process.
For time efficiency and full utilization of computation resources, we set the batch size as large as possible.
We implement gradient checkpointing in the finetuning process of $\text{CN-CLIP}\rm_{ViT\text{-}L/14@336px}$ and $\text{CN-CLIP}\rm_{ViT\text{-}H/14}$ for a larger batch size.
Table~\ref{tb:detailed_finetune} shows the specific settings of batch size, peaking learning rate, maximum epochs, and warmup iterations in the finetuning process.
We set other hyperparameters to be the same as those in pretraining by default.
We save the model parameters at the end of each epoch.
For MUGE, we report the best results on the validation set.
For Flickr30K-CN and COCO-CN, we choose the checkpoint with the best performance on the validation set and report the results on the test set.

\subsection{Cross-modal Retrieval with Longer Texts}
The results reported in Section~\ref{subsubsec:results} demonstrate the excellent cross-modal retrieval capability of Chinese CLIP.
Note that the average text lengths of MUGE, Flickr30K-CN, and COCO-CN are $7.4$, $19.7$, and $16.8$, respectively.
We also conduct finetuning experiments on the ICR~\citep{r2d2} dataset with an average text length of $45.3$. 
Experimental results are shown in Table~\ref{tb:ICR}.
Since the texts in the ICR dataset are longer, we set the maximum text length to $128$ for finetuning.
The results show that Chinese CLIP achieves state-of-the-art performance in cross-modal retrieval tasks with longer texts.


\subsection{Details About Experiments on Zero-shot Image Classification}

\begin{table*}[t]
\center
\small
\vskip 0.15in
\begin{adjustbox}{max width=1.\textwidth}
\begin{tabular}{@{\extracolsep{\fill}}lccc}
\toprule
Dataset & \#Labels & Test Size & Metric \\
\midrule
  Caltech-101~\citep{caltech} & 101 & 6,084 & Mean-per-class  \\
  CIFAR-10~\citep{cifar} & 10 & 10,000 &  Accuracy   \\
  CIFAR-100~\citep{cifar} & 100 & 10,000 & Accuracy   \\
  Country-211~\citep{clip} & 211 & 21,100 & Accuracy   \\
  DTD~\citep{dtd} & 47 & 1,880 & Accuracy  \\
  EuroSAT~\citep{eurosat} & 10 & 5,000 & Accuracy  \\
  FER-2013~\citep{clip} & 7 & 3,589 & Accuracy  \\
  FGVC-Aircraft~\citep{fgvc} & 100 & 3,333 & Mean-per-class \\
  Food-101~\citep{food101} & 101 & 25,250 & Accuracy \\
  GTSRB~\citep{gtsrb} & 43 & 12,630 & Accuracy  \\
  Hateful-Memes~\citep{hateful} & 2  & 500 & ROC AUC \\
  KITTI-Distance~\citep{kitti} & 4 & 711 & Accuracy  \\
  MNIST~\citep{mnist} & 10 & 10,000 & Accuracy \\
  Oxford Flowers-102~\citep{flowers} & 102 & 6,149 & Mean-per-class  \\
  Oxford-IIIT Pets~\cite{pets} & 37 & 3,669 & Mean-per-class \\
  PatchCamelyon~\citep{patchcamelyon} & 2 & 32,768 & Accuracy  \\
  Rendered-SST2~\citep{clip} & 2 & 1,821 & Accuracy  \\
  RESISC-45~\citep{resisc45} & 45 & 25,200 & Accuracy \\
  Stanford-Cars~\citep{stanford_cars} & 196 & 8,041 & Accuracy  \\
  Pascal VOC-2007~\citep{voc} & 20 & 4,952 & 11-point mAP \\
\bottomrule
\end{tabular}
\end{adjustbox}
\caption{Details of the image classification datasets in the ELEVATER benchmark. }
\label{tb:icinw}
\end{table*}

\begin{table*}[t]
\center
\small
\vskip 0.15in
\begin{adjustbox}{max width=1.\textwidth}
\begin{tabular}{@{\extracolsep{\fill}}lccccc}
\toprule
\multirow{2}{*}{Dataset} & CN-CLIP & CN-CLIP & CN-CLIP & CN-CLIP & CN-CLIP \\
 & $\rm_{RN50}$ & $\rm_{ViT\text{-}B/16}$ & $\rm_{ViT\text{-}L/14}$ & $\rm_{ViT\text{-}L/14@336px}$ & $\rm_{ViT\text{-}H/14}$ \\
\midrule
  Caltech-101  & 77.3 & 84.9 & 88.5 & 88.8 & 90.6  \\
  CIFAR-10  & 72.7 & 92.0 & 94.9 & 94.1 & 96.0  \\
  CIFAR-100   & 40.6 & 64.4 & 75.1 & 73.5 & 79.7  \\
  Country-211   & 7.7 & 15.2 & 21.0 & 25.4 & 25.3  \\
  DTD   & 36.9 & 43.6 & 44.2 & 43.8 & 51.2  \\
  EuroSAT   & 27.0 & 46.9 & 56.9 & 50.7 & 52.0  \\
  FER-2013   & 21.9 & 47.2 & 54.6 & 55.1 & 49.2  \\
  FGVC-Aircraft  & 5.4 & 12.8 & 16.0 & 17.1 & 26.2  \\
  Food-101   & 39.8 & 62.4 & 69.4 & 73.9 & 74.6  \\
  GTSRB   & 22.3 & 28.4 & 37.3 & 35.5 & 38.5  \\
  Hateful-Memes   & 50.3 & 56.2 & 53.4 & 52.8 & 54.7  \\
  KITTI-Distance  & 30.2 & 33.5 & 49.9 & 49.8 & 39.1  \\
  MNIST   & 50.2 & 67.6 &69.8 & 65.0 & 79.4  \\
  Oxford Flowers-102   & 30.7 & 52.2 &62.5 & 64.8 & 68.4  \\
  Oxford-IIIT Pets   & 48.7 & 73.0 &81.6 & 83.1 & 83.5  \\
  PatchCamelyon  & 47.7 & 54.0 & 63.5 & 62.9 & 52.4  \\
  Rendered-SST2  & 50.1 & 52.3 & 61.4 & 62.9 & 61.0  \\
  RESISC-45   & 49.3 & 58.7 & 65.2 & 65.8 & 66.9  \\
  Stanford-Cars  & 27.3 & 42.3 &49.8 & 54.1 & 71.8  \\
  VOC-2007  & 82.1 & 83.3 & 84.5 & 84.9 & 84.9  \\
  Average  & 40.9 & 53.5 & 60.0 & 60.2 & 62.3  \\
\bottomrule
\end{tabular}
\end{adjustbox}
\caption{Experimental results of the zero-shot image classification performance of models on ICinW. }
\label{tb:zs}
\end{table*}

We present the data statistics and metrics of the $20$ image classification datasets of the track ICinW in the ELEVATER benchmark in Table~\ref{tb:icinw}. 
For the adaptation of Chinese CLIP to the English-native benchmark, we apply a series of preprocessing strategies. Specifically, we translate the text descriptions of the labels and the templates for manual prompts to Chinese. 
For example, the labels in CIFAR-10 include ``car, dog, ...'', and we manually translate the words into Chinese. 
There are also particular cases, such as the labels in FGVC-Aircraft~\citep{fgvc}, which are difficult to translate or transliterate. 
We search the names on Google and figure out the best Chinese name for each label. 
Be that as it may, we cannot guarantee that we have the best Chinese translation, and more importantly, it is still hard for the Chinese pretrained model to understand some of the concepts, which may lead to unsatisfactory performance in the related downstream tasks. 
As to the templates, for some datasets, we use our translation of the templates provided by the ELEVATER toolkit,\footnote{\url{https://github.com/Computer-Vision-in-the-Wild/Elevater\_Toolkit\_IC}} and for the others, we use the translation of the templates from OpenAI CLIP.

We present the experimental results of all Chinese CLIP models on zero-shot image classification in Table~\ref{tb:zs}. 
It can be found that the scaling of model size can consistently bring improvements in model performance. 
The predictable improvements of scaling Chinese CLIP indicate that we can further scale up the model for better performance in the future work.  
However, it is still a pity that the tiny-size $\text{CN-CLIP}\rm_{RN50}$ saliently performs much worse than the ViT variants which are significantly larger. 
This shows that there is still much room for the small model to improve, and the knowledge transfer of CLIP from large models to small models should be an important research topic in multimodal representation learning. 

\subsection{Deployment}
\label{appendix:deployment}

Chinese CLIP is supported to be deployed into ONNX-based\footnote{\url{https://onnx.ai/}} and TensorRT-based\footnote{\url{https://developer.nvidia.com/tensorrt}} models, enabling faster text and vision representation generation (especially for online inference). In this section, we provide more details on the model conversion, as well as the performance improvement. 

Specifically, we employ the ONNX module in PyTorch with \textsc{onnxmltools}\footnote{\url{https://github.com/onnx/onnxmltools}} package to convert Chinese CLIP PyTorch models to ONNX-based models in FP16 precision. With the support of \textsc{onnxruntime-gpu}\footnote{\url{https://onnxruntime.ai/docs/install}} package, the ONNX-based models are able to infer on NVIDIA GPUs. The \textsc{TensorRT} package enables the TensorRT-based models obtained from ONNX-based models and provides the GPU inference runtime. Our TensorRT-based models are also in FP16 precision.

We benchmark the PyTorch implemented Chinese CLIP models with converted ONNX-based and TensorRT-based models using a server with a single NVIDIA T4 GPU. The server contains \num{16} Intel Xeon (Skylake) Platinum 8163 CPU cores with \num{64}GB memory. For each model, we inference the vision and text representations for $100$ batches and compute the average time. Simulating the scenario of online deployment, we use batch size of $1$. All the models infer with FP16 precision. Table~\ref{tb:deployment_speed} shows the comparisons of inference time cost. For almost all the model scales, ONNX-based and TensorRT-based models have optimized inference speed over native PyTorch implemented Chinese CLIP models, especially on smaller model sizes. For vision representation inference, the TensorRT-based models are around $1.3$ ($\text{CN-CLIP}\rm_{ViT\text{-}H/14}$) to $9.5$ ($\text{CN-CLIP}\rm_{RN50}$) times as fast as the Pytorch-based models. For text representation inference, the TensorRT-based models are around $6.2$ ($\text{CN-CLIP}\rm_{ViT\text{-}H/14}$) to $8.2$ ($\text{CN-CLIP}\rm_{ViT\text{-}L/14}$) times as fast as the PyTorch counterparts.

\begin{table*}[t]
\center
\small
\begin{adjustbox}{max width=1.\textwidth}
\begin{tabular}{@{\extracolsep{\fill}}lccccccc}
\toprule
  Inference Time per Sample (ms)
  &\multicolumn{3}{c}{Vision Representation}
  &\multicolumn{3}{c}{Text Representation}
  \\
\midrule
  Model Scale & PyTorch & ONNX & TensorRT & PyTorch & ONNX & TensorRT
  \\
\midrule
  $\text{CN-CLIP}\rm_{RN50}$
  & 12.93	& 5.04	& \textbf{1.36} & 	3.64 & 	0.95 &	\textbf{0.58}
  \\
  $\text{CN-CLIP}\rm_{ViT\text{-}B/16}$
  & 11.12	& 4.92	& \textbf{3.58}	& 12.47	& 3.42	& \textbf{1.54}
  \\
  $\text{CN-CLIP}\rm_{ViT\text{-}L/14}$
  & 21.19	& 17.10	& \textbf{13.08}	& 12.45	& 3.48	& \textbf{1.52}
  \\
  $\text{CN-CLIP}\rm_{ViT\text{-}L/14@336px}$
  & 47.11	& 48.40	& \textbf{31.59}	& 12.24	& 3.25	& \textbf{1.54}
  \\
  $\text{CN-CLIP}\rm_{ViT\text{-}H/14}$
  & 35.10	& 34.00	& \textbf{26.98}	& 23.98	& 6.01	& \textbf{3.89}
  \\  

\bottomrule
\end{tabular}
\end{adjustbox}
\caption{Inference speed comparisons among PyTorch, ONNX and TensorRT Chinese CLIP models.}
\label{tb:deployment_speed}
\end{table*}

We also evaluate the quality of ONNX-based and TensorRT-based model representations by measuring their zero-shot performance on MUGE retrieval dataset. Table~\ref{tb:deployment_muge} provides the experimental zero-shot results, which shows that the converted ONNX-based or TensorRT-based models keeps the quality of vision and text representations well, with no more than $0.1$ MR degradation in retrieval performance.

\begin{table*}[t]
\center
\small
\begin{adjustbox}{max width=1.\textwidth}
\begin{tabular}{@{\extracolsep{\fill}}lccccc}
\toprule
  \multirow{2}{*}{Model Scale} & \multirow{2}{*}{Framework}
  &\multicolumn{4}{c}{Zero-shot Performance}
  \\
   &  & R@1 & R@5 & R@10 & MR
  \\
\midrule
  \multirow{3}{*}{$\text{CN-CLIP}\rm_{RN50}$}
  & PyTorch	& 42.6	& 68.6	& 77.9	& 63.0 
  \\
  & ONNX	& 43.0 & 68.4 & 78.1 & 63.2
  \\
  & TensorRT	& 42.8 & 68.5 & 78.0 & 63.1
  \\ \midrule
  \multirow{3}{*}{$\text{CN-CLIP}\rm_{ViT\text{-}B/16}$}
  & PyTorch	& 52.1	& 76.7	& 84.4	& 71.1 
  \\
  & ONNX	& 52.0	& 76.8	& 84.3	& 71.1	
  \\
  & TensorRT	& 52.0	& 76.8	& 84.2	& 71.0
  \\ \midrule
  \multirow{3}{*}{$\text{CN-CLIP}\rm_{ViT\text{-}L/14}$}
  & PyTorch	& 56.3	& 79.8	& 86.2	& 74.1 
  \\
  & ONNX	& 56.4	& 80.0	& 86.3	& 74.2	
  \\
  & TensorRT	& 56.3 & 79.9	& 86.5	& 74.2
  \\ \midrule
  \multirow{3}{*}{$\text{CN-CLIP}\rm_{ViT\text{-}L/14@336px}$}
  & PyTorch	& 59.0	& 81.4	& 87.8	& 76.1
  \\
  & ONNX	& 59.2	& 81.4 & 87.6 & 76.1
  \\
  & TensorRT	& 59.2 & 81.7 & 87.5 & 76.1
  \\ \midrule
  \multirow{3}{*}{$\text{CN-CLIP}\rm_{ViT\text{-}H/14}$}
  & PyTorch	& 63.0	& 84.1	& 89.2	& 78.8 
  \\
  & ONNX	& 63.1	& 84.1	& 89.0	& 78.8	
  \\
  & TensorRT	& 63.1	& 84.2	& 89.1	& 78.8
  \\

\bottomrule
\end{tabular}
\end{adjustbox}
\caption{Zero-shot results on MUGE-Retrieval dataset among PyTorch, ONNX and TensorRT Chinese CLIP models.}
\label{tb:deployment_muge}
\end{table*}

\begin{figure*}[h] 
    \centering
    \includegraphics[width=1.0\linewidth]{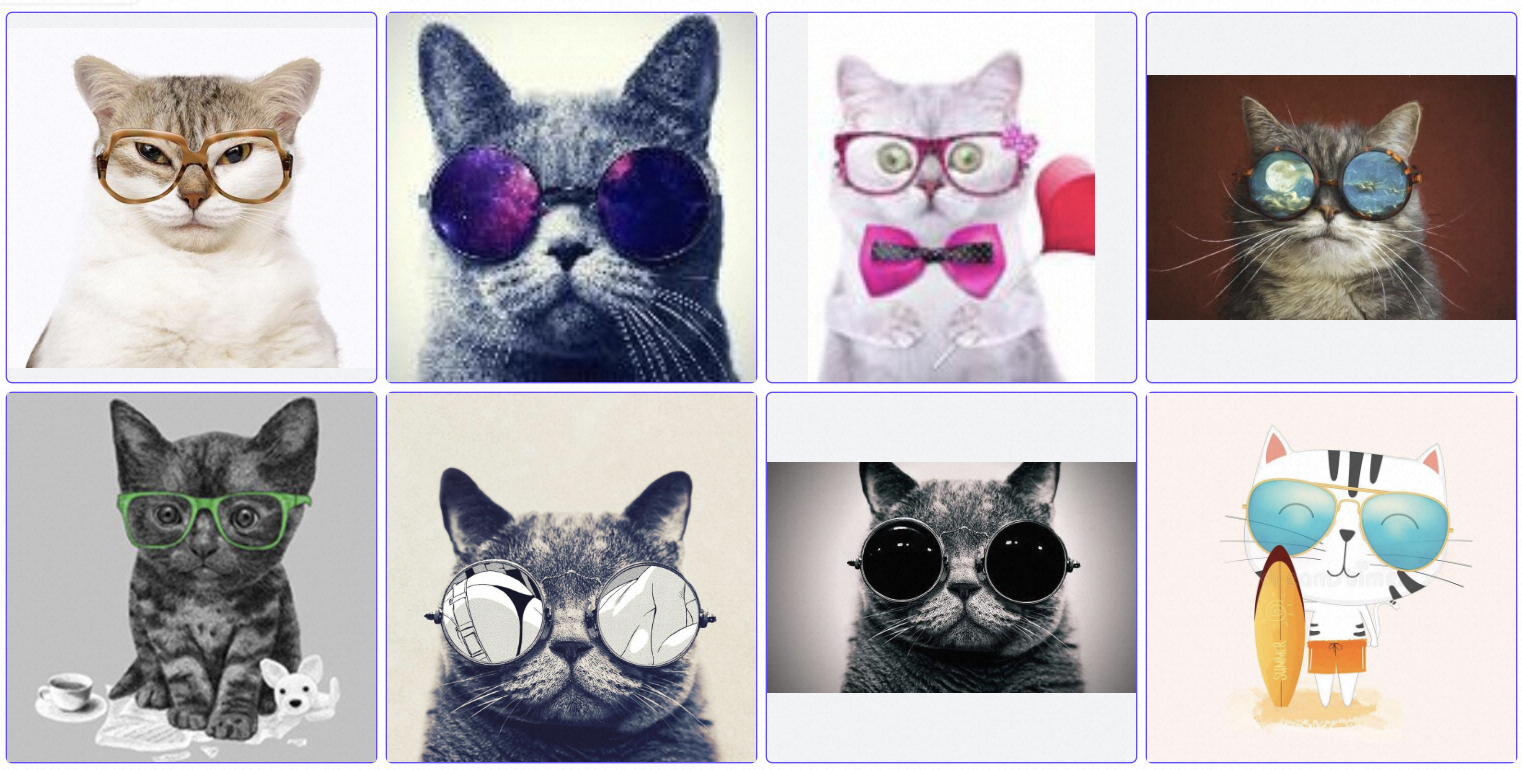}
    \caption{Retrieval results of the query ``a cat with glasses'' in Chinese.}
    \label{fig:demo_cat}
\end{figure*}

\begin{figure*}[h] 
    \centering
    \includegraphics[width=1.0\linewidth]{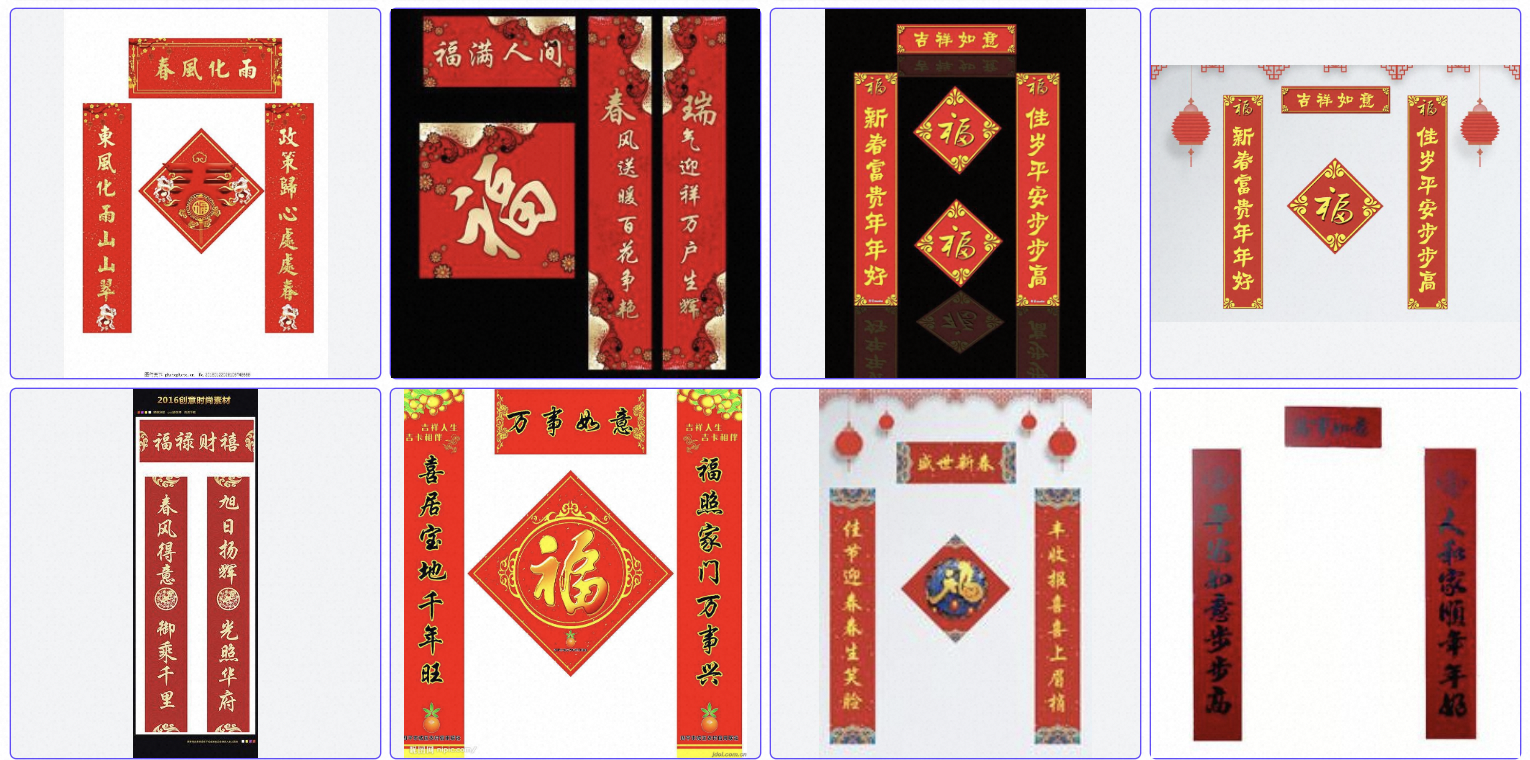}
    \caption{Retrieval results of the query ``Spring Festival couplet'' in Chinese.}
    \label{fig:demo_couplet}
\end{figure*}

\end{document}